\documentclass{article}

\usepackage{microtype}
\usepackage{graphicx}
\usepackage{subcaption}
\usepackage{booktabs} 

\usepackage{hyperref}

\newcommand\showfont{\expandafter\string\the\font}
\usepackage{enumitem}
\usepackage{multirow}
\usepackage{fvextra}
\DeclareUnicodeCharacter{03C0}{\ensuremath{\pi}} 

\usepackage[preprint]{icml2026}

\usepackage{amsmath}
\usepackage{amssymb}
\usepackage{mathtools}
\usepackage{amsthm}

\usepackage[capitalize,noabbrev]{cleveref}

\theoremstyle{plain}

\theoremstyle{definition}

\theoremstyle{remark}

\usepackage[textsize=tiny]{todonotes}

\icmltitlerunning{LinearizeLLM}

\begin{document}

\twocolumn[
  \icmltitle{LinearizeLLM: An Agent-Based Framework for LLM-Driven Exact Linear Reformulation of Nonlinear Optimization Problems}



  \icmlsetsymbol{equal}{*}

\begin{icmlauthorlist}
    \icmlauthor{Paul-Niklas Ken Kandora}{yyy}
    \icmlauthor{Simon Caspar Zeller}{yyy}
    \icmlauthor{Aaron Jeremias Elsing}{yyy}
    \icmlauthor{Elena Kuss}{sch}
    \icmlauthor{Steffen Rebennack}{yyy}
\end{icmlauthorlist}

\icmlaffiliation{yyy}{Institute for Operations Research, Karlsruhe Institute of Technology, Karlsruhe, Germany}
\icmlaffiliation{sch}{Institute for Information Systems, Reutlingen University, Reutlingen, Germany}

\icmlcorrespondingauthor{Paul-Niklas Kandora}{paul-niklas.kandora@kit.edu}

  \icmlkeywords{Machine Learning, ICML}

  \vskip 0.3in
]



\printAffiliationsAndNotice{}  

\begin{abstract}
Reformulating nonlinear optimization problems into solver-ready linear optimization problems is often necessary for practical applications, but the process is often manual and requires domain expertise. We propose LinearizeLLM, an agent-based LLM framework that produces solver-ready linear reformulations of nonlinear optimization problems. Agents first detect the nonlinearity pattern (e.g., bilinear products) and apply nonlinearity pattern-aware reformulation techniques, selecting the most suitable linearization technique. We benchmark on 40 instances: 27 derived from ComplexOR by injecting exactly-linearizable operators, and 13 automatically generated instances with deeply nested nonlinearities. LinearizeLLM achieves 73\% mean end-to-end overall success (OSR) across nonlinearity depths (8.3× higher than a one-shot LLM baseline; 4.3× higher than Pyomo). The results suggest that a set of pattern-specialized agents can automate linearization, supporting natural-language-based modeling of nonlinear optimization.
\end{abstract}

\section{Introduction}\label{sec:intro} 
Complex real-world optimization problems often involve nonlinear relationships between decision variables that make optimization problems computationally challenging to solve. A standard approach in Operations Research (OR) is to reformulate nonlinear optimization problems (NLOPs), transforming them into more tractable optimization problems \citep{williams2013model,vielma2015mixed}. These reformulations allow solvers such as Gurobi 12 \citep{gurobi2024} to apply LP/MILP algorithms on problem types, such as Linear Optimization Problems (LPs) or Mixed-Integer Linear Optimization Problems (MILPs). In practice, however, reformulating NLOPs into LPs typically requires expert knowledge. Practitioners without OR expertise often encounter intractable models, because they are unaware of the sophisticated reformulation techniques. This gap between advanced OR theory and what non-experts can readily apply limits broader adoption of optimization in industry and science \citep{chen2023mind}.

Meanwhile, LLMs may help reduce this expertise gap \citep{wasserkrug2025enhancing}. In particular, \citet{wasserkrug2025enhancing} highlight using LLMs to reformulate nonlinear optimization problems into linear ones as an open research direction. Recent LLM-based approaches have demonstrated the ability to perform complex reasoning tasks in OR. \citep{xiao2024chain} have begun to position LLMs as high-level translators of textual problem descriptions into formal optimization problems and even solve them with Chain-of-Experts. Additionally, OptiMUS introduced by \citep{ahmaditeshnizi2024optimus} is an LLM-driven framework that can formulate an MILP from a textual description, write corresponding solver code, and iteratively refine the optimization problem based on solution feedback. These systems have already shown that LLMs can alleviate the heavy dependence on domain experts in the modeling phase, making optimization tools more accessible to non-experts. However, a related challenge remains under-explored: Once an initial complicated NLOP is formulated from language, how can we reformulate that optimization problem into an equivalent linear model? In other words, beyond producing an optimization problem from language through LLM-based systems we ask whether an LLM-based system can also improve the model’s tractability by linearizing NLOPs? 

In this study we focus on nonlinear patterns that admit exact linear reformulations. Concretely, the current implementation supports exact reformulation techniques for absolute value, $\min/\max$, binary-continuous products, linear fractional terms (under sign conditions), and monotone objective transformations; see Appendix~\ref{app:recipes}. However, note that inexact linearization techniques, e.g., McCormick envelopes for purely continuous bilinear terms \citep{mccormick1976computability}, are also compatible with the agent architecture.
The above question of whether LLMs can be used to linearize problems is important because even an algebraically correct formulated optimization problem may be impractical to be solved if it involves nonlinear functions. In practice, many practitioners do not apply these techniques, and current LLM systems do not consistently perform this reformulation step. Recent evaluations highlight this gap \citep{wasserkrug2025enhancing}: When asked to reformulate a given optimization problem, an LLM often fails to produce an equivalent (or ``nearly'' equivalent) optimization problem. 

This paper presents \textit{LinearizeLLM}, an LLM-driven framework that reformulates a broad class of NLOPs into equivalent solver-ready LP/MILP models. We evaluate reformulations using a solver-based suite that combines end-to-end objective preservation with stage-wise success metrics for detection, reformulation validity, and compilation; we summarize this combined criterion as the \textit{Overall Success Rate (OSR)}. Since multiple optima and auxiliary-variables make “argmin equality” subtle to certify, we do not directly compare solution-sets but additionally report argmin comparisons in Appendix~\ref{app:depth_1_results} as additional empirical evidence that the obtained solutions are consistent across formulations up to renaming and auxiliaries.
Specifically, as our main contributions, we
\begin{enumerate}[label=(\roman*)]
    \item propose an agent architecture in which each nonlinear term is handled by a specialized \textit{reformulation agent} instructed to derive an exact linearization pattern after reading in the original problem formulation in \LaTeX{} code; the agents then coordinate to assemble a solver-ready model;
    \item release a benchmark dataset of 40 nonlinear optimization problems consisting of 27 manually constructed instances obtained by injecting exactly-linearizable nonlinearity patterns into selected ComplexOR problems \citep{xiao2024chain}, and 13 automatically generated instances featuring highly nested nonlinearities for evaluating robustness under deeper nesting;
    \item conduct an empirical evaluation of \textit{LinearizeLLM} on the dataset, benchmarking end-to-end overall success across nonlinearity nesting depths, and comparing against both a one-shot LLM reformulator and a deterministic Pyomo-based linearization pipeline. We further assess robustness under controlled \LaTeX{} perturbations of the optimization problem and perform a context-ablation study to quantify how structured metadata affects reformulation reliability.
\end{enumerate}

\section{Mixed-Integer Nonlinear Programming}
Mixed-Integer Nonlinear Problems (MINLPs) appear in many real-world applications, including process systems engineering, energy operations, logistics, and finance, where discrete decisions must be made under nonlinear dynamics \citep{Belotti2013ActaNumerica}.

Formally, a general MINLP can be written as:

\begin{align*}
\min_{x \in \mathbb{R}^n,\, y \in \mathbb{Z}^m} \quad & f(x, y) \\
\text{s.t.} \quad & g_i(x, y) = 0, \quad \forall i \\
                  & h_j(x, y) \leq 0, \quad \forall j
\end{align*}

where $f(x, y)$, $g_i(x, y)$ and $h_j(x,
y)$ are potentially nonlinear functions in $(x,
y)$, and $y$ contains integer-valued decision variables. If all functions are linear, the problem is a MILP; if, in addition, there are no $y$ variables, i.e., $m=0$, then the problem is an LP.

\section{Motivation}
First, one practical limitation is solver support for a broad class of nested nonlinearities. While modern NLOP/MINLP solvers accept broad classes of nonlinear terms, certain nested constructs, such as compositions involving monotone transformations and piecewise linear functions either violate standard differentiability assumptions or fall outside supported operator sets in common solver interfaces \citep{waechter2006ipopt}. As a result, such models often require an explicit exact reformulation before they can be reliably handed to a solver such as Gurobi or open-source engines like HiGHS \citep{highs2023} and CBC \citep{cbc2022}.

Modern MILP solvers incorporate presolve routines and even automatic detection of certain nonlinear patterns. However, these built-in features are limited and difficult to audit. By contrast, an explicit reformulation handled outside the solver supports transparency and auditability. The user (or model auditor) can inspect the introduced auxiliary variables and linear constraints, verifying that they correctly represent the original nonlinear relations. The auditability supports review: the reformulated model is human-readable and can be double-checked line by line, unlike solver-internal transformations that are hidden from view. Auditability is important in high-stakes applications where one must ensure the reformulation has not altered the problem’s intent or feasibility region.

\textit{LinearizeLLM} (i) generates a fully documented set of auxiliary variables and constraints, giving auditors line-by-line traceability; (ii) outputs its model in standard algebraic form, so the same file can be fed to any LP/MILP engine or embedded as a linear sub-problem inside decomposition schemes; and (iii) employs pattern-specialized LLM agents that recognize and linearize nonlinear (nested) patterns beyond the cases typically handled by solver presolve. In doing so, \textit{LinearizeLLM} encodes reformulation rules into an automated, transparent, and portable pipeline, enabling practitioners to use LP/MILP solvers on models that would otherwise require nonlinear optimization support.

\section{Related Work}

\subsection{LLMs for Model Formulation}
Early work on natural language to optimization focused on the modeling task, i.e., formulating MILPs from problems described with natural language.
\citep{pmlr-v220-ramamonjison23a} introduced NL4Opt, a publicly available dataset and NeurIPS 2022 competition with the task of translating real-world problems into LPs. Several other studies later focused on this problem. \citep{li2023synthesizing} developed a three-phase framework to formulate MILPs from natural text and extended the NL4Opt dataset to evaluate their approach. Additional work on this task includes OptiMUS, introduced by \citep{ahmaditeshnizi2024optimus}, an LLM based agent, that models natural language problems as MILPs, writes and evaluates the solver-ready code. \citep{xiao2024chain} is closest to our work. The authors introduced a multi-agent framework \textit{Chain-of-Experts}, where each agent is assigned to a specific task. Their framework is capable of generating solver-ready code for OR problems and is evaluated on the new \textit{ComplexOR} dataset. 

However, these approaches assume linearity and therefore do not address nonlinear formulations.
In the OptimAI approach \citep{thind2025optimai}, a suitable solver is selected based on the structure and requirements of the given optimization problem. However, no reformulation of the problem is performed, and the problem is solved in its original mathematical form.
A recent study by \citep{wasserkrug2025enhancing} evaluated ChatGPT’s \citep{openai} ability to perform algebraic reformulations. When asked to replace a nonlinear absolute-value constraint with linear constraints (a common linearization task), ChatGPT produced an equivalent formulation in convex cases. 
However, for non-convex cases that require integer auxiliary variables, the model’s answers were often incomplete; it tended to omit the necessary binary variables, yielding incorrect formulations. This suggests that LLMs can recognize certain reformulation patterns (indeed, ChatGPT \textit{knew} that “max” constraints can be rewritten as linear inequalities) but may fail to enforce logical consistency unless explicitly guided.

These results motivate explicit guidance or agent-based decomposition when exact linearization is required. Motivated by this gap, we present \textit{LinearizeLLM}, a multi-agent pipeline that reformulates each detected nonlinear pattern individually to produce solver-ready linear models. The next section details its workflow.

\section{LinearizeLLM Workflow}
\textit{LinearizeLLM} transforms an NLOP into an equivalent LP/MILP in three stages: a \textit{detection agent} parses the original NLOP and reports each nonlinear pattern; a deterministic structural policy partitions detected nonlinearity patterns into distinct nonlinear terms and determines the reformulation order; and the loop repeats until no nonlinear patterns remain after being processed by the \textit{reformulation agents}, yielding a linear optimization problem.
We consider a fixed set of exactly linearizable patterns: absolute value, $\min/\max$, binary--continuous products, linear fractional terms (with denominator sign constraints), and monotone objective transformations. Reformulation recipes and required preconditions are given in Appendix~\ref{app:recipes}.

\subsection{Detection agent}
The \textit{detection agent} is an LLM-based agent tasked for identifying and mapping the hierarchy of nonlinearities within a given optimization problem. The agent performs a scan to distinguish between actual decision variables and constant parameters. The agent detects nonlinear patterns that can be reformulated exactly into linear or mixed-integer linear forms; if no such patterns are detected, the workflow terminates.

\textbf{Parameter-aware Contextualization.}\ 
To minimize false positives, where linear terms are incorrectly identified as nonlinear (e.g., $x \cdot p$, where $p$ is a constant parameter and $x$ is a decision variable), the agent is provided with a parameter context that explicitly defines the set of decision variables. The agent is instructed to ignore terms involving only parameters or constants, so that mathematically linear forms are excluded from the results.

\textbf{Hierarchical Pattern-Scanning.}\ 
The agent is prompted to reason about the \textit{nesting structure} of nonlinearities. For each detected term \(\pi \in \Pi\), the agent reports its mathematical expression, its specific nonlinearity type \(\tau(\pi)\), and its nesting depth \(d(\pi)\). A top-level nonlinearity (one not contained within another nonlinear operator) is assigned a depth of $1$, while inner nonlinearities have increasing depth values. 

The agent is instructed to output a dependency structure for each $\pi$ by tracking \textit{parent} and \textit{child} node. In this context, a parent node is an outer nonlinear operator (e.g., a square root in the objective function) whose argument contains another nonlinear expression, which is then defined as its child node (e.g., a bilinear product nested inside that square root). Nested terms can then be decomposed recursively and replaced by auxiliary variables to maintain mathematical equivalence of the optimization problems. We can see a simple example for this structure in Figure~\ref{fig:dependency_tree}.

\begin{figure}[ht]
\centering
\begin{tikzpicture}[
    level distance=1.8cm,
    edge from parent/.style={draw, -latex, thick},
    every node/.style={draw, rounded corners, rectangle, fill=gray!5, align=center, minimum height=1.1cm, minimum width=3cm, font=\small}
]

\node {Parent: $\sqrt{\cdot}$ \\ (Depth 1)}
    child {
        node {Child: $x \cdot y$ \\ (Depth 2)}
    };

\begin{scope}[every node/.style={font=\footnotesize\itshape, text=gray}]
    \draw [dashed, gray!40] (2.2,0) -- (4,0) node[right] {Top-level operator};
    \draw [dashed, gray!40] (2.2,-1.8) -- (4,-1.8) node[right] {Nested nonlinearity};
\end{scope}

\end{tikzpicture}
\caption{Dependency tree for the nested expression $\sqrt{x \cdot y}$. This hierarchy illustrates how the agent tracks the relationship between operators, ensuring that the inner child node (the bilinear product) is resolved before the outer parent node (the monotone transformation).}
\label{fig:dependency_tree}
\end{figure}
\subsection{Structural Policy}
The \textit{structural policy} is a deterministic sequencing mechanism that defines the order in which detected nonlinear terms \(\Pi\) are processed. Unlike the detection agent, which relies on the semantic reasoning of an LLM, the structural policy applies a fixed rule based on the dependency structure of the (nested) nonlinear patterns constructed during detection.

\textbf{Depth-Based Sequencing.}\ 
To preserve equivalence of the model during reformulation, nested nonlinearities must be resolved such that any child term is linearized and replaced by an auxiliary variable before its parent term is processed. The policy achieves this by sorting all terms \(\pi \in \Pi\) in descending order of their nesting depth \(d(\pi)\). This ensures a bottom-up traversal of the dependency structure, where the most nested terms are handled first.

\textbf{Why bottom-up matters.}\
Consider $\max\{0,xy\}$ with bilinear (continuous-binary) child $xy$: linearizing the outer $\max$ first via a binary selector $z$ forces activation of the $xy$ branch only when $z=1$, introducing continuous--binary bilinearities such as $z\cdot(xy)$. Processing bottom-up instead reformulates $xy$ into an auxiliary variable $w$ first, after which $\max\{0,w\}$ is piecewise-linear and can be linearized without creating new nonlinear terms.

\textbf{State Consistency.}\ 
By enforcing this specific sequence, the policy ensures that at any iteration \(k\), the specialized reformulation agent \(\Phi_{\tau}\) receives a model state \(p^{(k-1)}\) where the arguments of the current term \(\pi_k\) have already been simplified into linear forms. This reduces the instruction complexity and limits growth of nested expressions.

\subsection{Reformulation agent}
While the detection agent identifies nonlinear patterns, the actual transformation of these patterns is performed by the \textit{type-specific reformulation agents} \(\{\Phi_{\tau}\}\). Rather than relying on static reformulation templates, the system invokes an agent specialized for the specific nonlinearity type \(\tau\) (e.g., bilinear). This ensures that each transformation is handled by an expert module tailored to the unique mathematical requirements of that specific pattern.

\textbf{General reformulation instructions}\ 
The agent is guided by a universal instruction that emphasizes producing a linear LP/MILP and preserving equivalence between the original NLOP and the resulting LP/MILP. This generalized approach allows the system to handle the supported nonlinearities; from standard bilinear products to complex monotone transformations, without requiring a separate, specialized agent for every possible mathematical form. This agent operates on general principles of optimization theory rather than fixed "lookup tables."

\textbf{Iterative Reformulation.}\ 
The reformulation process is iterative: at each step \(k\), the specialized agent \(\Phi_{\tau}\) replaces the nonlinear term \(\pi_k\) in the current optimization problem \(p^{(k-1)}\) with a linear auxiliary variable and introduces the corresponding linear constraints. The agent receives the nonlinear term’s exact location in the model, along with the relevant problem parameters, to support accurate reformulation. An important feature of these agents is their ability to provide "post-reformulation" instructions, for example, when simplifying an objective function through a monotone transformation, the agent provides the exact reconstruction formula (e.g., recovering the original value from a squared objective). Each step is documented within the evolving LaTeX model, ensuring that the final linear formulation \(\hat{p}\) is mathematically transparent and ready for solver execution.

\subsection{The Workflow of LinearizeLLM}\label{sec:LinearizeLLM-workflow}

Let \(\mathcal{P}\) be the set of solvable nonlinear optimization problems. For any \(p\in\mathcal{P}\) the original problem instance is denoted by \(p \in \mathcal{P}\). We denote by \(\Theta\) the set that contains all parameter sets \(\theta_p\), \(\forall p\). The parameter set \(\theta_p\) collects all parameters (e.g., index sets) associated with problem \(p\) and is assumed to remain invariant with respect to the reformulation.

Each problem \(p\in\mathcal{P}\) may contain nonlinear patterns, potentially in nested form. We collect all detected nonlinear patterns in a set \(\Pi\), where each \(\pi \in \Pi\) represents a (nested) nonlinear pattern and is annotated with (i) a nonlinearity type \(\tau(\pi) \in \mathcal{T}\) (e.g., bilinear, absolute value, etc.), (ii) its location in the model (e.g. constraint or objective), and (iii) its nesting depth \(d(\pi) \in \mathbb{N}\). To represent nested structures, we store for each term a parent node \(\mathrm{parent}(\pi) \in \Pi \cup \{\text{None}\}\) and a set of children nodes \(\mathrm{children}(\pi) \subseteq \Pi\).

We define a \emph{detection agent} as the following mapping
\[
  \mathrm{Detect} : \mathcal{P} \times \Theta \to 2^{\Pi},
\]
which is called once on the original problem and returns the set of all nonlinear terms \(\Pi := \mathrm{Detect}(p, \theta_p)\). If \(\Pi = \emptyset\), the problem is linear and the workflow terminates.

To ensure that children nodes are linearized before their parents, we define a \emph{structural policy} that orders the terms in \(\Pi\) by their nesting depth. This results in an ordered sequence of terms \((\pi_1, \dots, \pi_{|\Pi|})\) such that \(d(\pi_k) \geq d(\pi_{k+1})\) for all \(k\).

We employ a \emph{generalized reformulation agent} \(\{\Phi_{\tau} \mid \tau \in \mathcal{T}\}\). The reformulation agent
\[
  \Phi_{\tau} : \mathcal{P} \times \Theta \times \Pi \to \mathcal{P}
\]
is instructed to derive exact linear formulations based on the nonlinearity pattern $\tau$ as well as the problem $p^{(k-1)}$ and the context $\theta_p$ the agent receives. The reformulation process is viewed as a sequence of intermediate problems \((p^{(k)})_{k=0}^{|\Pi|}\), where \(p^{(0)} = p\) is the original model and, for each \(k=1,\dots,|\Pi|\),
\[
  p^{(k)} := \Phi_{\tau(\pi_k)}\!\bigl(p^{(k-1)}, \theta_p, \pi_k\bigr).
\]
In this sequence, at each iteration \(k\), the agent \(\Phi_{\tau(\pi_k)}\) replaces the nonlinear term \(\pi_k\) in the current model \(p^{(k-1)}\) with its exact linear reformulation, yielding the updated model \(p^{(k)}\). The final reformulated linear model is then \(\hat p := p^{(|\Pi|)}\).

\begin{algorithm}[H]
  \caption{\textit{LinearizeLLM} Structure-Driven Workflow}
  \label{alg:linllm-main-loop}
  \begin{algorithmic}[1]
    \REQUIRE Initial problem \(p\), parameter set \(\theta_p\)
    \STATE \(\Pi \leftarrow \mathrm{Detect}(p, \theta_p)\)
    \STATE \((\pi_1, \dots, \pi_{|\Pi|}) \leftarrow \mathrm{SortByDepthDescending}(\Pi)\)
    \STATE \(p^{(0)} \leftarrow p\)
    \FOR{\(k = 1, \dots, |\Pi|\)}
        \STATE \(\tau \leftarrow \mathrm{Type}(\pi_k)\)
        \STATE \(p^{(k)} \leftarrow \Phi_{\tau}(p^{(k-1)}, \theta_p, \pi_k)\) 
    \ENDFOR
    \STATE \(\hat p \leftarrow p^{(|\Pi|)}\)
    \STATE \textbf{return} \(\hat p\)
  \end{algorithmic}
\end{algorithm}
The workflow operates as follows. The detection agent identifies the set of all nonlinear terms \(\Pi\) within the original problem \(p\) using the context \(\theta_p\). A structural policy sorts these terms by their nesting depth in descending order to establish a bottom-up processing sequence. The ordering procedure is performed by $\mathrm{SortByDepthDescending}(\cdot)$. \(p^{(0)}\) as the initial nonlinear optimization problem is initialized for the iterative reformulation process. The algorithm iterates through each sorted nonlinear term \(\pi_k\) in the sequence. The specific nonlinearity type \(\tau\) is identified for the current term. The reformulation agent \(\Phi_{\tau}\) linearizes the term \(\pi_k\) within the current problem state, replacing it with an exact linear equivalent and necessary auxiliary variables. Once all terms are processed, the final state \(p^{(|\Pi|)}\) is captured as the fully linearized model \(\hat{p}\). The resulting linear formulation is returned for subsequent code generation and optimization.

\begin{figure}[ht]
    \centering
    \begin{tikzpicture}[>=stealth, node distance=1.5cm]
    
    \node (p0) at (1.5,0.25) [align=center] {$p^{(0)}$};
    \coordinate (a) at (2,0.25);
    \coordinate (b) at (3,0.25);
    \draw[->, thick] (a) -- (b);
    \node[anchor=center] at (2.5, -1.2) {\includegraphics[width=1cm]{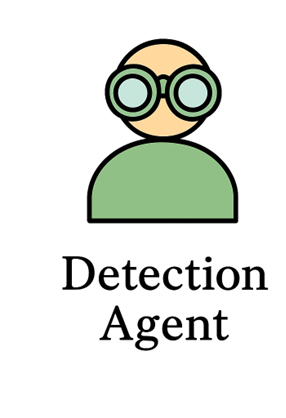}};
    \node (pi) at (3.5 ,0.25) {$\Pi$};
    
    \coordinate (c) at (4,0.25);
    \coordinate (d) at (4.5,0.25);
    
    \draw[->, thick] (c) -- (d);
    
    \node (pi_sort) at (5.5, 0.25) [font=\small]{$(\pi_1, \dots, \pi_{|\Pi|})$};
    
    \coordinate (e) at (6.5,0.25);
    \coordinate (f) at (6.75,0.25);
    
    \draw[-, thick] (e) -- (f);

    \coordinate (g) at (6.75,2);
    \coordinate (g2) at (6.75,1);
    \coordinate (g3) at (6.75, -0.2);
    
    \draw[-, thick] (f) -- (g);
    \draw[-, thick] (f) -- (g3);
    \coordinate (h) at (7,2);
    \coordinate (h2) at (7,1);
    \coordinate (h3) at (7,-0.2);
    
    \node (p1) at (7.5,2) [align=center] {$p^{(1)}$};
    \node (p2) at (7.5,1) [align=center] {$p^{(2)}$};
    \draw[->, thick] (p1) -- (p2);
    
    \node (dots) at (7.5, 0.5) {$\vdots$};
    
    \node (p) at (7.5,-0.2) [align=center] {$p^{(|\Pi|)}$};
    
    \node (p_hut) at (7.5,-0.2) [align=center] {$p^{(|\Pi|)}$};
    \node (p) at (8.5, -0.2) [align=center] {$\rightarrow \hat{p} $};
    
    \draw[->, thick] (g) -- (h);
    \draw[->, thick] (g2) -- (h2);
    \draw[->, thick] (g3) -- (h3);

    \coordinate (u) at (2,-1);
    \coordinate (c) at (3.1 ,-1);
    

    \definecolor{agentgreen}{HTML}{A3C793}
    \definecolor{agentorange}{HTML}{F4A07A}
    
    \node[anchor=center] at (7.5, -1.2) {\includegraphics[width=1cm]{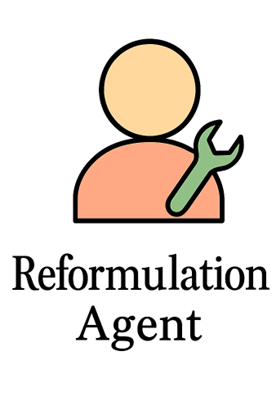}};
    
    
    \end{tikzpicture}
    \caption{Illustration of the \textit{LinearizeLLM} workflow. We initialize \textit{LinearizeLLM} with the NLOP/MINLP $p^{(0)}$ and terminate with the LP/MILP reformulation $\hat{p}$.}
\end{figure}

\section{Experiments}
\subsection{ComplexOR-NL dataset}
Our \textit{ComplexOR-NL} dataset is derived from the ComplexOR dataset by \citep{xiao2024chain}. We created 40 nonlinear instances by injecting exactly linearizable nonlinearities into selected base instances. In total, the benchmark comprises 27 manually constructed instances based on ComplexOR and 13 automatically generated instances designed to include more deeply nested and syntactically diverse compositions of nonlinearities. Each instance is feasible, has a finite optimum and admits an exact linear reformulation.
The dataset is designed to both cover realistic optimization use-cases, obtained by augmenting real-world–motivated ComplexOR problems, and to evaluate baselines and \textit{LinearizeLLM} with nested nonlinearities through automatically generated instances. Table~\ref{tab:problem_instances} summarizes how frequently each nonlinearity appears across the 40 instances (counted once per file).

\begin{table}[ht]
    \centering
    \begin{tabular}{@{}l c@{}}
        \toprule
        \textbf{Nonlinearity}          & \textbf{Count} \\
        \midrule
        Bilinear terms                  & 18 \\
        Min operator                    & 26 \\
        Max operator                    & 21 \\
        Absolute-value                  & 24 \\
        Linear fractional               & 8 \\
        Monotone transformations        & 5 \\
        \bottomrule
    \end{tabular}
    \caption{Number of \textsc{ComplexOR-NL} problem files (out of 40) that contain each nonlinear construct at least once; multiple occurrences within the same file (e.g., two max terms) are counted only once.} \label{tab:problem_instances}
\end{table}


\subsection{Performance Metrics}\label{subsec:metrics}
Rather than relying on manual inspection of reformulations, which is difficult to make exhaustive and reproducible, we evaluate \textit{LinearizeLLM} primarily using \textbf{Overall Success Rate (OSR)}, which measures whether the pipeline produces a solver-ready LP/MILP whose optimal objective value (after any stated objective mappings) matches that of the original nonlinear model within a numerical tolerance. For each instance, we build a reference model and solve it with Gurobi Version 12 using either hand-written (hand-crafted benchmarks) or auto-generated (synthetic benchmarks) Python code, applying exact reformulations into Gurobi-supported primitives (including general constraints and bounded big-M for binary–continuous products, with required domain checks). For additional diagnostics, we report three component metrics that localize errors to (i) nonlinearity detection (DSR), (ii) validity of the produced LP/MILP as judged by solver status and classification (RSR), and (iii) compilation of the \LaTeX{} model into executable solver code (CSR). We do not provide a formal proof of argmin-set equivalence. Since (MI)LPs often admit multiple optima and our reformulations introduce auxiliary variables, exact equality of the solver-returned minimizer vectors is not a well-posed requirement. Instead, OSR provides a practical end-to-end check for optimal-value preservation. Appendix~\ref{app:depth_1_results} additionally reports side-by-side optimal solutions (objective values and variable assignments) for both the original and linearized models; in OSR-successful cases, these traces are typically consistent after accounting for auxiliary variables and renaming, providing additional evidence for solution equivalence. DSR/RSR/CSR help attribute failures to specific stages of the workflow:

\textbf{Detection Success Rate (DSR):} This metric measures the proportion of instances in which the correct nonlinearity pattern is successfully identified by the \textit{detection agents}. 

\textbf{Reformulation Success Rate (RSR):} This metric assesses whether the \textit{reformulation agents} were able to successfully produce a valid LP/MILP formulation (model is not classified as infeasible or unbounded). 

\textbf{Compiler Success Rate (CSR):} This metric captures the success rate of compiling the \LaTeX{}-formulated optimization problems into executable code. 

\textbf{Overall Success Rate (OSR):} Our overall performance measure, OSR denotes the proportion of total runs that are free of detection, reformulation, or compilation errors and yield a reformulated model whose optimal objective value, after mapping back to the original objective when monotone transformations are applied, matches (with a tolerance of $\epsilon=10^{-4}$) that of the original nonlinear problem, $p^{(0)}$, as solved by Gurobi. A tolerance of $10^{-4}$ accounts for numerical precision differences introduced during reformulation, such as auxiliary variables, while still validating optimal-value agreement of the reformulated model with the original formulation on the evaluated instances. In addition to the other metrics, OSR therefore tests that the linearized model attains the original optimum value, which is necessary but not sufficient for equivalence; Appendix~\ref{app:depth_1_results} provides complementary solution comparisons as additional empirical evidence beyond objective-value matching.

Each experiment is evaluated over three independent runs; we report averages. 

\subsection{Baselines}\label{subsec:base}
We compare \textit{LinearizeLLM} against the following baselines.

\textbf{One-shot LLM.}
We evaluate the necessity of our multi-agent orchestration by introducing a one-shot baseline using the same LLM as used for the agents within \textit{LinearizeLLM}. This agent is tasked with simultaneous detection and exact reformulation in a single LLM-based agent. 

\textbf{Deterministic rules-based linearizer.}
We compare against a deterministic baseline using \texttt{pyomo.gdp}, a rule-based reformulation engine. We translate the \LaTeX\ model into a Generalized Disjunctive Programming (GDP) representation and apply Pyomo’s standard reformulations to obtain a MILP, providing a reproducible non-LLM reference.

We report DSR for \textit{LinearizeLLM} and the one-shot baseline, since these methods generate an explicit pattern set that can be scored against the ground-truth nonlinearities. In contrast, the pyomo baseline is a transformation-based pipeline without a separable detection stage (the model is directly encoded into GDP constructs), so DSR is not well-defined; we therefore report only RSR/CSR/OSR for this baseline. For pyomo, we flag reformulation success when the instance is successfully encoded as a valid GDP model and Pyomo’s standard GDP-to-MILP transformation applies without error.

We provide detailed descriptions for our baselines in Appendix~\ref{app:baselines}.

\subsection{Experimental Setup}
We evaluate our pipeline (Algorithm~\ref{alg:linllm-main-loop}), which converts optimization problems formulated in \LaTeX{} into solver-ready Python and executes them with Gurobi (\texttt{gurobipy}). Unless stated otherwise, LLM agents use Gemini~2.5~Flash with \texttt{temperature}\,=\,0.05. Reported metrics are averaged over 3 random seeds. Prompt templates and exact reformulation recipes are provided in Appendix~\ref{app:prompts} and~\ref{app:recipes}, respectively.

\section{Results}
\subsection{Head-to-Head baseline comparison}

\begin{table*}[ht]
\centering
\small
\setlength{\tabcolsep}{5pt}
\begin{tabular}{lcccccccccccc}
\toprule
& \multicolumn{12}{c}{\textbf{Nonlinearity Composition Depth}} \\
\cmidrule(lr){2-13}
\textbf{Method}
& \multicolumn{4}{c}{\textbf{$d = 1$}}
& \multicolumn{4}{c}{\textbf{$d = 2$}}
& \multicolumn{4}{c}{\textbf{$d \geq 3$}} \\
\cmidrule(lr){2-5} \cmidrule(lr){6-9} \cmidrule(lr){10-13}
& DSR & RSR & CSR & OSR
& DSR & RSR & CSR & OSR
& DSR & RSR & CSR & OSR \\
\midrule
LinearizeLLM
& \textbf{89.47} & \textbf{100} & \textbf{100} & \textbf{78.95}
& \textbf{96.97} & \textbf{84.85} & \textbf{96.97} & \textbf{66.67}
& \textbf{96.67} & \textbf{100.00} & \textbf{100.00} & \textbf{73.33} \\
One-shot LLM
& 12.28 & 96.49 & 100 & 10.53
& 12.12 & 69.70 & 100.00 & 9.09
& 6.67 & 90.00 & 100.00 & 6.67 \\
Pyomo
& -- & 29.82 & 29.82 & 22.81
& -- & 30.3 & 30.3 & 15.15
& -- & 16.67 & 16.67 & 13.33 \\
\bottomrule
\end{tabular}
\vspace{0.5em}
\caption{Head-to-head baseline comparison across nonlinearity composition depths. We report Detection Success Rate (DSR), Reformulation Success Rate (RSR), Compiler Success Rate (CSR), and Overall Success Rate (OSR), where OSR measures end-to-end optimal objective-value preservation (within tolerance) after any stated objective mappings. Results are averaged over 3 runs.}
\label{tab:baseline-comparison-depth}
\end{table*}

Table~\ref{tab:baseline-comparison-depth} reports a head-to-head comparison of \textit{LinearizeLLM} against the baselines using the success metrics from Section~\ref{subsec:metrics}. A comprehensive breakdown of computational efficiency, including per-agent inference times and token consumption across all nesting depths, is provided in Appendix~\ref{app:costs}. 

The results in Table~\ref{tab:baseline-comparison-depth} demonstrate a clear difference in performance, with \textit{LinearizeLLM} outperforming both the one-shot and the Pyomo approach across all complexity depths.

\textit{LinearizeLLM} maintains consistently high detection accuracy ($\sim$90--97\%) across nesting depths, whereas the one-shot baseline degrades sharply ($\sim$6--12\%). This gap is consistent with the possibility that in a one-shot setup, pattern identification competes with reformulation and code generation for limited model capacity and attention.

We additionally observe that \textit{LinearizeLLM} exhibits higher Reformulation Success Rate (RSR) than Detection Success Rate (DSR) for $d=1$. Mathematically, this appears counter-intuitive; however, it indicates a self-correction capability within the pipeline. Even if the detection agent fails to explicitly label a nonlinearity pattern (leading to a DSR failure), the subsequent reformulation and code generation agent are still provided with the full \LaTeX{} optimization problem. Because these agents re-evaluate the mathematics while generating logic, they may recognize and linearize the missed nonlinear pattern. This indicates that the multi-stage pipeline can tolerate some early-stage errors.

The Pyomo approach shows a distinct profile where RSR and CSR are identical (e.g., 29.8\% at $d=1$). This indicates that the primary bottleneck for problem reformulation appears to be syntactic. If the LLM successfully navigates the complex Pyomo syntax (CSR), the underlying Pyomo transformation routines usually succeed (RSR). The subsequent drop to OSR (22.8\%) suggests that even when the code runs, the LLM occasionally maps the objective function or constraints incorrectly.

\textit{LinearizeLLM} is stable across depths, maintaining an OSR of 73.33\% even at $d \geq 3$. In contrast, both the One-shot and Pyomo methods struggle as the ``nesting'' of nonlinearities increases. The 100\% RSR for \textit{LinearizeLLM} at $d=1$ and $d \geq 3$ suggests that once the system correctly identifies the problem structure, the implemented reformulation rules are reliable on this benchmark.

\subsection{Robustness to Input Perturbations}
We test robustness to syntactic variation in \LaTeX{} by perturbing each instance at five levels: \textbf{L0 (clean)}, \textbf{L1 (format noise)}, \textbf{L2 (macros/refactoring)}, \textbf{L3 (structure-preserving reordering)}, and \textbf{L4 (combined)}. We evaluate 5 randomly selected instances per depth category ($d=1$, $d=2$, $d\geq 3$). Formal perturbation definitions and examples are given in Appendix~\ref{app:perturbations}.

\begin{figure*}[ht]
  \centering
  \includegraphics[]{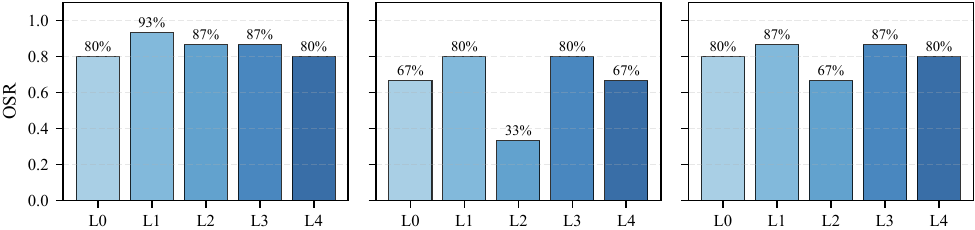}
  \caption{OSR across perturbation levels (L0-L4) for three groups of nonlinear instances, stratified by nesting depth: \textbf{left} shows problems with nested depth $d=1$, \textbf{middle} shows $d=2$, and \textbf{right} shows $d\ge 3$.}
  \label{fig:osr-robustness}
\end{figure*}
In Figure~\ref{fig:osr-robustness}, OSR is similar under pure formatting noise (L1) and structure-preserving reordering (L3) across all depth groups, indicating that the pipeline is insensitive to superficial \LaTeX{} variation. The main degradation occurs under L2 (macro + refactor), most notably for $d=2$ where OSR drops to $33\%$. Since these perturbations preserve the full problem semantics, this indicates that surface-form refactoring (macro aliases and operator rewrites) is challenging for end-to-end compilation, and can be more disruptive than constraint reordering. Even under the combined setting (L4), OSR does not consistently degrade (and returns close to the clean baseline for $d=1$ and $d\ge 3$), suggesting that failures are associated with specific refactoring transformations rather than perturbation density alone.

\subsection{Context ablation study}\label{sec:context_ablation}
We ablate the structured metadata provided to the agents to measure how context affects end-to-end reformulation success. We consider six settings: \textbf{C1 (Minimal)} provides only the \LaTeX{} model (objective/constraint type only); \textbf{C2 (Variables Only)} adds decision variable definitions; \textbf{C3 (Parameters Only)} provides parameter names; \textbf{C4 (Vars + Param Names)} combines C2 and C3; \textbf{C5 (Vars + Param Values)} additionally provides numerical parameter values; and \textbf{C6 (Full Context)} further includes local constraint information. We focus on DSR and OSR to isolate the effect of context on (i) correctly identifying nonlinear patterns and (ii) achieving end-to-end solver-verified execution.

We evaluate all settings on 10 randomly sampled instances (5 with $d=2$ and 5 with $d\ge 3$). Full details concerning the study are given in Appendix~\ref{app:context_ablation}.

\begin{table}[ht]
\centering
\small
\setlength{\tabcolsep}{6pt}
\begin{tabular}{lcc}
\toprule
\textbf{Context setting} & \textbf{Depth = 2} & \textbf{Depth $\geq$ 3} \\
\midrule
C1: None & 0.00\% & 0.00\% \\
C2: Vars only & 0.00\% & 0.00\% \\
C3: Params only & 0.00\% & 0.00\% \\
C4: Vars + Params & 0.00\% & 0.00\% \\
C5: Vars + Params & 86.67\% & 33.33\% \\
C6: Full & 73.33\% & 46.67\% \\
\bottomrule
\end{tabular}
\caption{Context ablation on compositional instances: Overall Success Rate (OSR) across depth categories.}
\label{tab:ablation_osr}
\end{table}

\begin{table}[ht]
\centering
\small
\setlength{\tabcolsep}{6pt}
\begin{tabular}{lcc}
\toprule
\textbf{Context setting} & \textbf{Depth = 2} & \textbf{Depth $\geq$ 3} \\
\midrule
C1: None & 100.00\% & 93.33\% \\
C2: Vars only & 100.00\% & 100.00\% \\
C3: Params only & 100.00\% & 93.33\% \\
C4: Vars + Params & 100.00\% & 86.67\% \\
C5: Vars + Params & 100.00\% & 86.67\% \\
C6: Full & 100.00\% & 93.33\% \\
\bottomrule
\end{tabular}
\caption{Context ablation on compositional instances: Detection Success Rate (DSR) across depth categories.}
\label{tab:ablation_dsr}
\end{table}

According to Table~\ref{tab:ablation_osr}, the main observation is 0\% OSR when concrete parameter values are withheld, regardless of other metadata. Even with full knowledge of variable names and indexing (C4), the agents do not achieve OSR success without the numerical values (C5). This suggests that providing parameter values is important for generating executable reformulations in this setting. In Depth 2 instances from Table~\ref{tab:ablation_dsr}, detection (DSR) remains a perfect 100\% even with zero context (C1), suggesting that these nonlinearities are easier to detect from surface form. However, in Depth 3 instances, the DSR fluctuates and even drops when metadata is incomplete (C4/C5). This indicates that as nesting depth increases, the agent's ability to even correctly identify the nonlinear terms depend in parts on its understanding of the model's underlying structure.

Overall, this study demonstrates that while syntactic awareness is sufficient for basic term identification, numerical grounding and semantic localization are the important prerequisites for successful end-to-end optimization reformulation.

\paragraph{Limitations.}
\textit{LinearizeLLM} currently handles a fixed set of nonlinear patterns with known exact LP/MILP reformulations (Appendix~\ref{app:recipes}), and does not aim to linearize general nonlinear functions beyond these cases or to obtain (tight) relaxations when exact reformulations are not possible. Our main automatic metric (OSR) checks optimal objective-value agreement (within tolerance) between the original and reformulated models under a solver run; it does not certify argmin-set equivalence. Results are based on 40 instances and one LLM and could differ for other problem instances or LLMs.

\section{Conclusion}
We introduced \textit{LinearizeLLM}, an agent-based framework for producing exact LP/MILP reformulations of nonlinear optimization problems written in \LaTeX{}. \textit{LinearizeLLM} decomposes reformulation into (i) nonlinearity detection with explicit nesting structure, (ii) a deterministic depth-based policy that enforces bottom-up processing, and (iii) pattern-specialized reformulation agents that generate linear models.

We released a benchmark dataset of 40 nonlinear optimization problems that admit exact linearization, combining manually constructed variants of ComplexOR problems with automatically generated, deeply nested instances. Across different levels of nesting-depths, \textit{LinearizeLLM} outperforms both a one-shot LLM reformulator and a deterministic Pyomo-based pipeline in overall success rate, with high stage-wise success rates. Robustness experiments show that the pipeline is insensitive to superficial \LaTeX{} changes, with the main sensitivity arising from macro-based refactoring. Finally, a context ablation study highlights that numerical grounding and structured metadata is important for reliable end-to-end reformulation on more compositional instances.

These results suggest that pattern-specialized LLM agents, orchestrated by simple structural rules, can reduce the gap between nonlinear modeling convenience and linear-solver tractability.

\section*{Impact Statement}


This paper presents work whose goal is to advance the field of Machine
Learning. There are many potential societal consequences of our work, none which we feel must be specifically highlighted here.



\bibliographystyle{icml2026}
\bibliography{icml2026}

@inproceedings{
xiao2024chain,
title={Chain-of-Experts: When {LLM}s Meet Complex Operations Research Problems},
author={Ziyang Xiao and Dongxiang Zhang and Yangjun Wu and Lilin Xu and Yuan Jessica Wang and Xiongwei Han and Xiaojin Fu and Tao Zhong and Jia Zeng and Mingli Song and Gang Chen},
booktitle={The Twelfth International Conference on Learning Representations},
year={2024},
url={https://openreview.net/forum?id=HobyL1B9CZ}
}

@article{thind2025optimai,
  title={OptimAI: Optimization from Natural Language Using LLM-Powered AI Agents},
  author={Thind, Raghav and Sun, Youran and Liang, Ling and Yang, Haizhao},
  journal={arXiv preprint arXiv:2504.16918},
  year={2025}
}

@article{mccormick1976computability,
  title={Computability of global solutions to factorable nonconvex programs: Part I—Convex underestimating problems},
  author={McCormick, Garth P},
  journal={Mathematical programming},
  volume={10},
  number={1},
  pages={147--175},
  year={1976},
  publisher={Springer}
}

@InProceedings{ahmaditeshnizi2024optimus,
  title = 	 {{O}pti{MUS}: Scalable Optimization Modeling with ({MI}){LP} Solvers and Large Language Models},
  author =       {Ahmadi{T}eshnizi, Ali and Gao, Wenzhi and Udell, Madeleine},
  booktitle = 	 {Proceedings of the 41st International Conference on Machine Learning},
  pages = 	 {577--596},
  year = 	 {2024},
  editor = 	 {Salakhutdinov, Ruslan and Kolter, Zico and Heller, Katherine and Weller, Adrian and Oliver, Nuria and Scarlett, Jonathan and Berkenkamp, Felix},
  volume = 	 {235},
  series = 	 {Proceedings of Machine Learning Research},
  month = 	 {21--27 Jul},
  publisher =    {PMLR},
  url = 	 {https://proceedings.mlr.press/v235/ahmaditeshnizi24a.html}
}

@misc{gurobi2024,
  author       = {{Gurobi Optimization, LLC}},
  title        = {Gurobi Optimizer},
  year         = {2024},
  howpublished = {\url{https://www.gurobi.com}},
  note         = {Accessed: 2025-07-14}
}

@misc{openai,
  author       = {{OpenAI}},
  title        = {Introducing {ChatGPT}},
  year         = {2022},
  month        = nov,
  day          = {30},
  howpublished = {\url{https://openai.com/index/chatgpt/}},
  note         = {Accessed: 2026-01-28}
}

@manual{cbc2022,
  author        = {Forrest, John and Ralphs, Ted K.},
  title         = {{CBC} (COIN-OR Branch and Cut) 2.10.7 User Guide},
  organization  = {COIN-OR Foundation},
  year          = {2022},
  url           = {https://github.com/coin-or/Cbc}
}

@article{vielma2015mixed,
  author  = {Vielma, Juan Pablo},
  title   = {Mixed-Integer Linear Programming Formulation Techniques},
  journal = {SIAM Review},
  volume  = {57},
  number  = {1},
  pages   = {3--57},
  year    = {2015},
  doi     = {10.1137/130915303}
}

@article{chen2023mind,
  title={Mind the gap between research and practice in operations management},
  author={Chen, Xiaohong and Deng, Tianhu and Shen, Zuo-Jun Max and Yu, Yi},
  journal={IISE Transactions},
  volume={55},
  number={1},
  pages={32--42},
  year={2023},
  publisher={Taylor \& Francis}
}

@InProceedings{pmlr-v220-ramamonjison23a,
  title = 	 {NL4Opt Competition: Formulating Optimization Problems Based on Their Natural Language Descriptions},
  author =       {Ramamonjison, Rindranirina and Yu, Timothy and Li, Raymond and Li, Haley and Carenini, Giuseppe and Ghaddar, Bissan and He, Shiqi and Mostajabdaveh, Mahdi and Banitalebi-Dehkordi, Amin and Zhou, Zirui and Zhang, Yong},
  booktitle = 	 {Proceedings of the NeurIPS 2022 Competitions Track},
  pages = 	 {189--203},
  year = 	 {2022},
  editor = 	 {Ciccone, Marco and Stolovitzky, Gustavo and Albrecht, Jacob},
  volume = 	 {220},
  series = 	 {Proceedings of Machine Learning Research},
  month = 	 {28 Nov--09 Dec},
  publisher =    {PMLR},
  url = 	 {https://proceedings.mlr.press/v220/ramamonjison23a.html}
}

@article{li2023synthesizing,
  title={Synthesizing mixed-integer linear programming models from natural language descriptions},
  author={Li, Qingyang and Zhang, Lele and Mak-Hau, Vicky},
  journal={arXiv preprint arXiv:2311.15271},
  year={2023}
}

@article{wasserkrug2025enhancing, 
title={Enhancing Decision Making Through the Integration of Large Language Models and Operations Research Optimization}, volume={39}, url={https://ojs.aaai.org/index.php/AAAI/article/view/35090}, DOI={10.1609/aaai.v39i27.35090}, abstractNote={Many critical business and societal decisions in areas such as supply chain and healthcare involve numerous potential actions, complex constraints, and goals that can be modeled as objective functions. Mathematical optimization, a core area in Operations Research (OR), provides robust, mathematically grounded methodologies to address such decisions and has shown tremendous benefits in many applications. However, its application requires the creation of accurate and efficient optimization models, necessitating rare expertise and considerable time, creating a barrier to widespread adoption in decision-making. Thus, it is a long-standing goal to make these capabilities widely accessible. The advent of Large Language Models (LLMs) has made advanced Artificial Intelligence (AI) capabilities widely accessible through natural language. LLMs can accelerate expert work in creating formal models like computer programs, and emerging research indicates they can also speed up the development of optimization models by OR experts. We, therefore, propose integrating and advancing LLM and optimization modeling to empower organizational decision-makers to model and solve such complex problems without requiring deep expertise in optimization. In this work, we present our vision for democratizing optimization modeling for organizational decision-making by such a combination of LLMs and optimization modeling. We identify a set of fundamental requirements for the vision’s implementation and describe the state of the art through a literature survey and some experimentation. We show that a) LLMs already provide substantial novel capabilities relevant to realizing this vision, but that b) major research challenges remain to be addressed. We also propose possible research directions to overcome these gaps. We would like this work to serve as a call to action to bring together the LLM and OR optimization modeling communities to pursue this vision, thereby enabling much more widespread improved decision-making and increasing by orders of magnitude the benefits AI and OR can bring to enterprises and society.}, number={27}, journal={Proceedings of the AAAI Conference on Artificial Intelligence}, author={Wasserkrug, Segev and Boussioux, Leonard and den Hertog, Dick and Mirzazadeh, Farzaneh and Birbil, {{\c{S}}.} {\.I}lker and Kurtz, Jannis and Maragno, Donato}, year={2025}, month={Apr.}, pages={28643-28650} }

@article{Belotti2013ActaNumerica,
  author  = {Pietro Belotti and Christian Kirches and Sven Leyffer and Jeff Linderoth and James Luedtke and Ashutosh Mahajan},
  title   = {Mixed-integer nonlinear optimization},
  journal = {Acta Numerica},
  volume  = {22},
  pages   = {1--131},
  year    = {2013},
  doi     = {10.1017/S0962492913000032},
  url     = {https://www.cambridge.org/core/journals/acta-numerica/article/mixedinteger-nonlinear-optimization/2D0CE8CDA53363A31ADE8689565517BD}
}

@article{hart2011pyomo,
  author  = {Hart, William E. and Watson, Jean-Paul and Woodruff, David L.},
  title   = {Pyomo: modeling and solving mathematical programs in Python},
  journal = {Mathematical Programming Computation},
  year    = {2011},
  volume  = {3},
  number  = {3},
  pages   = {219--260},
  doi     = {10.1007/s12532-011-0026-8},
  url     = {https://doi.org/10.1007/s12532-011-0026-8}
}

@article{highs2023,
	author = {Huangfu, Q. and Hall, J. A. J.},
	journal = {Mathematical Programming Computation},
	number = {1},
	pages = {119--142},
	title = {Parallelizing the dual revised simplex method},
	volume = {10},
	year = {2018}}

@article{waechter2006ipopt,
  title   = {On the implementation of an interior-point filter line-search algorithm for large-scale nonlinear programming},
  author  = {W{\"a}chter, Andreas and Biegler, Lorenz T.},
  journal = {Mathematical Programming},
  volume  = {106},
  number  = {1},
  pages   = {25--57},
  year    = {2006},
  doi     = {10.1007/s10107-004-0559-y},
  url     = {https://link.springer.com/article/10.1007/s10107-004-0559-y}
}

@book{williams2013model,
  author    = {Williams, H. Paul},
  title     = {Model Building in Mathematical Programming},
  edition   = {5},
  year      = {2013},
  publisher = {John Wiley \& Sons},
  isbn      = {978-1-118-44333-0}
}

\newpage
\appendix
\onecolumn
\section{Baselines}\label{app:baselines}
\subsection{One-shot LLM}
The one-shot baseline replaces the orchestrated \textit{LinearizeLLM} workflow with a single LLM agent. This agent receives the full \LaTeX\ optimization problem and is instructed to provide the full exact linear reformulation in a single response. The resulting reformulation is passed to the standard code-generation agent to produce executable Python code. The full instruction templates for this agent are provided in Section~\ref{app:one_shot_prompt}.

\subsection{Deterministic rules-based linearizer}
As a deterministic benchmark, we utilize the \texttt{pyomo.gdp} module, a specialized engine within the Pyomo ecosystem designed for logic-based optimization problem transformations. Because this engine requires models to be defined in a specific ``logical form,'' our pipeline first performs symbolic preprocessing via \texttt{SymPy} to normalize nonlinear expressions. These are then translated into Generalized Disjunctive Programming (GDP) constructs, specifically disjuncts and logical constraints. We then invoke Pyomo’s internal transformation suite to apply exact linear reformulations \cite{hart2011pyomo}. This baseline is particularly strong for logical and piecewise-linear nonlinearities, though it is inherently limited to patterns supported by the \texttt{pyomo.gdp} layer. It serves as a check on whether LLM-based reformulation can compare against the precision of established symbolic optimization tools. The full instruction templates for this agent are provided in Section~\ref{app:pyomo_prompt}.

\section{Prompt templates}\label{app:prompts}
This section contains all the prompts used in the \textit{LinearizeLLM} for pattern detection, reformulation, and code generation. The original markdown prompt files have been transformed into pure \LaTeX{} code with proper mathematical formatting and structured documentation. The content is the same. 

\subsection{Detection Prompt}
\begin{Verbatim}[breaklines=true,breakanywhere=true]
You are the DETECT agent. Given a LaTeX optimization problem, find **all nonlinear terms** Your goal is to identify everything that prevents the model from having a clearly maintained linear structure (MILP/LP).

**CRITICAL:** If a term (e.g., a square root, product, or quotient) involves only parameters/constants, it is a constant coefficient and should be ignored. Similarly, terms that are mathematically linear (e.g., a variable divided by a constant parameter) must not be flagged. **Do not return or mention these linear terms.**

For each detected term π, return:
- term_id: unique id
- expression: the nonlinear expression (as written)
- type: one of {{bilinear, minmax, absolute_value, quotient, monotone_transformation, affine, other}}
- location:
    - component: objective OR constraint
    - constraint_id: index/name if component=constraint
    - sense: one of {{<=, >=, =}} if component=constraint
    - lhs_text: full LHS text of the constraint (if component=constraint)
    - rhs_text: full RHS text of the constraint (if component=constraint)
    - side: "lhs" if term is on LHS, "rhs" if on RHS, "objective" if in objective
    - rhs: the RHS expression/token (e.g., nrhs_31) if component=constraint
    - needs_value: true if the exact value of the term must be represented (objective, equality, or reused), false otherwise.
- depth d(π): 0 for top-level; increase with nesting
- parent_id: id of parent term if nested, else null
- children_ids: ids of direct children (including affine branches)
- pattern_role: semantic role of the node:
    - "definition": sense is "=" and node covers almost full LHS or RHS
    - "lhs_upper_bounded": node is on LHS and sense is <=
    - "lhs_lower_bounded": node is on LHS and sense is >=
    - "rhs_upper_bounding_expr": node is on RHS and sense is <=
    - "rhs_lower_bounding_expr": node is on RHS and sense is >=
    - "embedded": otherwise
- notes: brief clue if type=other

**DAG Construction Rules:**
- Construct a complete expression DAG.
- Every operator node (min, max, abs, etc.) MUST list ALL of its arguments as children.
- If an argument is a linear/affine expression (e.g., a^\top x + b), represent it as a lightweight node:
    - type: "affine"
    - children_ids: []
- This ensures arity is preserved (e.g., a min{{...}} with 4 branches has 4 children).

What counts as nonlinear (non-exhaustive):
- Products of different **decision variables** (bilinear) such as x*y where x and y are decision variables.
- Min/Max, absolute value involving decision variables.
- Quotients and divisions where the **denominator contains decision variables**.
- Monotone transforms of decision variables or linear forms: sqrt(\cdot), log(\cdot), exp(\cdot), positive powers, other general smooth monotone functions.
- Any nested combination of the nonlinear patterns mentioned above (e.g., sqrt(max_i |...|), log(sum of products), etc.).

What counts as a decision variable:
- Decision variables are symbols that appear in the outer optimization goal operator, i.e., below the main operator $\min/\max$ (e.g., $\min_{{x}}$ or $\min_{{x,y}}$), and are chosen by the optimizer to minimize/maximize the objective.
- In contrast, $\max(\cdot)$/$\min(\cdot)$ patterns that occur **within** the objective or constraints are part of the optimization problem's structure (nonlinear patterns) and do not introduce additional decision variables, unless they are replaced by auxiliary (epigraph) variables in a reformulation.
- **CRITICAL**: If we have something like $\min_{{x}}$ where $x$ represents a vector/array, and within the objective or constraints $x$ appears indexed such as $x_{{ij}}$, then each indexed component (e.g., $x_{{ij}}$) is still a decision variable (a component of $x$).

Requirements:
- Cover every nonlinear term involving decision variables; include intermediate nested subterms IF and ONLY IF they are also nonlinear.
- Differentiate clearly: ignore and **do not return** expressions that mathematically just contain parameters and no decision variables.
- Preserve nesting: children_ids capture the inner nonlinear subterms; depth increases inward.
- If a subterm is linear (e.g., the argument of a sqrt(\cdot) or exp(\cdot) is a linear sum), it is NOT a nonlinear subterm and should NOT be returned as a separate term.
- If no nonlinearities exist, return an empty list (has_nonlinearities=false).

Input:
- LaTeX model: {latex_model}
- Parameter context: {parameter_context}

Output format:
{format_instructions}

\end{Verbatim}

\subsection{Reformulation Prompt}
\begin{Verbatim}[breaklines=true,breakanywhere=true]
You are an expert in exact linearization of nonlinear terms for mixed-integer/linear optimization.

Task:
- Reformulate the given nonlinear pattern(s) exactly into a linear (MILP/LP) form. Use the most compact linear formulation of the nonlinear pattern in terms of variables and constraints.
- **SURGICAL REPLACEMENT RULE (CRITICAL):** Identify the exact `pattern_text` within the model. Replace **ONLY** this specific occurrence with a new, unique auxiliary variable (e.g., aux_1). Do **NOT** reformulate other nonlinear parts of the model that do not match the current `pattern_text`. Do **NOT** apply global transformations (like Charnes-Cooper) unless the `pattern_text` itself is the root of that transformation.
- **BOTTOM-UP REFORMULATION CONTEXT:** You are processing a nested tree of nonlinearities from the leaves up. The `full LaTeX model` may already contain auxiliary variables (e.g., aux_1) from deeper reformulations. Treat these as standard linear variables.
- **ROOT-LEVEL OPTIMIZATION:** If the `location context` indicates this term is the root of an objective or a constraint, you may reformulate the model structure directly (e.g., removing a monotone square root from the objective) instead of using an auxiliary, if it leads to a more efficient MILP.
- Clearly maintain a linear structure in the updated optimization problem; ensure all terms, constraints, and the objective function are linear after reformulation.
- If you introduce new variables, the resulting constraints and objective MUST be linear in those variables.
- Keep the full LaTeX model consistent; return the **ENTIRE** updated model including all original linear parts and all added auxiliary variables and constraints. **DO NOT truncate the model or return only the reformulated part.**
- The ultimate goal is to find linear optimization problems (LPs/MILPs) that are equivalent to the nonlinear optimization problems (NLPs) handed-over by the user.
- Both formulations (the LP/MILP and NLP versions) should have the same optimal points and therefore optimal objective values.

**Pattern-Specific Requirements:**
- **MONOTONE TRANSFORMATION:** When simplifying the objective by removing a monotone function (e.g., \( \min \sqrt{{f(x)}} \)), you MUST provide the exact Python expression required to reconstruct the original objective value. Enter this in the `post_processing` field (e.g., "math.sqrt(model.ObjVal)").

Input:
- pattern_type: {pattern_type}
- pattern_text: {pattern_text}
- full LaTeX model to update:
{latex_model}
- location context (where the term appears):
{location_info}
- parameter info (if provided, for validation only):
{param_info}

Output requirements:
{format_instructions}

Notes:
- If Big-M is used, derive and state M explicitly.
- Preserve semantic equivalence; do not approximate.


\end{Verbatim}

\subsection{One-shot Prompt}\label{app:one_shot_prompt}
\begin{Verbatim}[breaklines=true,breakanywhere=true]
You are an Operations Research expert specialized in **exact linear reformulations** of nonlinear patterns in optimization models.

## TASK

From the given LaTeX model and parameter context:

1. Detect all linearizable nonlinear patterns.
2. Reformulate each nonlinear pattern you detect **exactly** as a linear/mixed-integer linear problem LP/MILP using the tightest valid technique.
3. Utilize the best reformulation strategy in order to end-up with a fully linear optimization problem (LP/MILP).
4. Output a **single fully linear LaTeX model** that is mathematically equivalent.

## INPUT
- **LaTeX model:** {latex_model}
- **Parameter context:** {parameter_context}
- **Parameter values:** {param_info}
- **Decision variables:** {sets_info}

## PATTERNS
Please check for nonlinearity patterns that can be linearized exact and apply appropriate reformulation techniques on those patterns.

## OUTPUT FORMAT

**DETECTION**
- Nonlinearities: YES/NO
- Patterns: List specific patterns found, grouped by type
  - If no patterns found, write "NONE"
  - If patterns found, list them like: "Bilinear: x*y, Min: min(x,y), Max: max(x,y)"

**REFORMULATION REPORT**
For each detected pattern family:
- Pattern: expression and indices
- Technique: chosen method
- Bounds / M: numeric or "n/a"
- Aux vars: new variables
- Verification: one-line justification
- For monotone transformations: apply the transformation in the code after deriving the optimal objective function value

**REFORMULATED MODEL (LaTeX)**
- Full linearized model with:
  - Decision variables (including new ones)
  - Linear objective
  - Original constraints
  - New reformulation constraints

**REFORMULATION INFORMATION**
- Concise summary of transformations applied, bounds, Big-M values, and domain conditions.

**POST-CODE HANDOVER (MONOTONE)**
- Apply: YES/NO
- Direction: increasing/decreasing
- Targets: expressions/variables to transform (list)
- Bounds/domain: required bounds or domain conditions
- Transformation note: brief formula or mapping to apply after code generation

**FINAL CHECKS**
- Residual nonlinearities: NONE (or list if any)
- Big-M constants numeric and tightened: YES/NO
- Index sets consistent and defined: YES/NO

## RULES
- Output must be LP/MILP only.
- Stop at first technique that is Applicable=YES and Exact=YES.
- Focus only on mathematical reformulation - code generation will be handled separately.
\end{Verbatim}

\subsection{Pyomo Conversion Prompt}\label{app:pyomo_prompt}
\begin{Verbatim}[breaklines=true,breakanywhere=true]
You are an expert in Operations Research and Mixed-Integer Optimization. Convert the provided LaTeX-formulated optimization model into efficient and immediately executable Python code using Gurobi's gurobipy API.

TASK REQUIREMENTS:
- Accurately translate the LaTeX model into gurobipy Python code.
- Maintain complete mathematical equivalence between LaTeX and Python.
- Clearly maintain a linear structure in the generated code; ensure that all reformulated terms are represented using linear variables and constraints.
- Clearly define all parameters, variables, constraints, and objectives.
- Integrate the provided parameters explicitly and correctly.
- Ensure the Python code is clear, efficient, and ready to run without modification.

PROVIDED INFORMATION:

PARAMETER CONTEXT:
{parameter_context}

AVAILABLE PARAMETERS:
{param_info}

{sets_info}

REFORMULATION INFORMATION:
{reformulation_context}

LATEX MODEL:
{latex_model}

GUIDELINES FOR CODE GENERATION:
- Start with imports:
from gurobipy import *

- Define parameters with provided numeric values.
- Map abstract indices and summations from LaTeX to concrete Python loops or efficient vectorized expressions.
- Use gurobipy methods effectively:
  - model.addVars() for indexed variables.
  - model.addVar() for single variables.
  - quicksum() for efficient summations.

- Clearly comment reformulations if applicable (e.g., "# McCormick envelopes for bilinear terms").
- Handle parameter dictionaries safely (e.g., param.get(key, default)).

- Finish your script with:
model.optimize()

CRITICAL QUICKSUM RULES:
- NEVER use conditional expressions inside quicksum() like: quicksum(x[i] for i in range(n) if condition)
- Instead, use explicit loops with LinExpr or separate quicksum calls
- For conditional summations, use this pattern:
  ```python
  # WRONG: quicksum(x[i] for i in range(n) if condition[i])
  # CORRECT:
  expr = LinExpr()
  for i in range(n):
      if condition[i]:
          expr += x[i]
  ```
- Or use separate quicksum calls for different conditions

IMPORTANT:
- Do NOT include markdown formatting or fences.
- Output should be only executable Python code with explanatory comments where necessary.

Generate ONLY executable Python code without any formatting, explanations, or markdown: 
\end{Verbatim}

\section{Exact Linearization Recipes}\label{app:recipes}

\noindent
We collect compact examples showing how to derive exact LP/MILP counterparts for the nonlinear patterns used in the paper.

\subsection{Bilinear products}

\paragraph{Case A: binary $\times$ binary.}
Let $b_1,b_2\in\{0,1\}$ and introduce $w=b_1 b_2$. An exact linearization is
\begin{align}
  w &\le b_1, \qquad
  w \le b_2, \qquad
  w \ge b_1 + b_2 - 1,\qquad
  0 \le w \le 1.
\end{align}
(Integrality of $w$ is implied when $b_1,b_2$ are binary.)

\paragraph{Case B: binary $\times$ bounded continuous.}
Let $b\in\{0,1\}$, $x\in[L,U]$ with known bounds, and $z=b\,x$.
Then the following four inequalities are \emph{exact}:
\begin{align}
  z \le U b,\qquad
  z \ge L b,\qquad
  z \le x - L(1-b),\qquad
  z \ge x - U(1-b).
\end{align}
When $b=0$ these force $z=0$; when $b=1$ they force $z=x$.

\subsection{Minimum and maximum of linear functions}

\paragraph{Constraint splitting for $\min$.}
For linear functions $f_k(x)$, $k\in\mathcal{K}$, the constraint
\[
  t \le \min_{k\in\mathcal{K}} \{ f_k(x) \}
\]
is equivalent to the set of linear constraints $t \le f_k(x)\ \forall k\in\mathcal{K}$.

\paragraph{Epigraph for $\max$ (objective).}
To solve $\min_x \ \max_{k\in\mathcal{K}} f_k(x)$ introduce an auxiliary $z$ and use
\begin{align}
  \min_{x,z}\ \ z \quad \text{s.t.}\quad z \ge f_k(x)\ \ \forall k\in\mathcal{K}.
\end{align}
This epigraph encoding is the standard exact reformulation when $\max$ appears in the objective.

\subsection{Absolute value}

\paragraph{Objective form.}
To minimize $|t|$ (with $t$ linear), introduce $y$ and write
\begin{align}
  \min\ y \quad \text{s.t.}\quad y \ge t,\ \ y \ge -t .
\end{align}

\paragraph{Equality form.}
To enforce $y = |t|$ exactly without binaries, use the positive/negative parts:
\begin{align}
  t &= t^+ - t^-, \qquad y = t^+ + t^-, \qquad t^+,t^- \ge 0 .
\end{align}

\subsection{Linear fractional (Charnes--Cooper)}

Consider the fractional objective
\[
  \min_x\ \frac{a^\top x + b}{c^\top x + d}
  \quad\text{s.t.}\quad F x \le g,\ \ c^\top x + d > 0.
\]
Charnes--Cooper sets $t=\frac{1}{c^\top x + d}>0$, $y = x t$, giving the LP
\begin{align}
  \min_{y,t}\ \ & a^\top y + b\,t \\
  \text{s.t.}\ \ & F y \le g\,t,\qquad c^\top y + d\,t = 1,\qquad t \ge 0.
\end{align}
(Contrast with a \emph{partial} cross-multiplication $u(c^\top x+d)\!\ge\!a^\top x+b$,
which leaves the bilinear term $u\,x$ and is therefore nonlinear.)

\subsection{Monotone transformations}

Let $\phi:\mathbb{R}\to\mathbb{R}$ be strictly monotone.

\paragraph{Objectives.}
If $\phi$ is strictly increasing, then $\min\ \phi(g(x))$ is equivalent to $\min\ g(x)$
(solve in $g$ and, if needed, report back the original value via $\phi$).
Example: minimizing $\exp(s)$ is equivalent to minimizing $s$ (and reporting $\exp(s^\star)$).

If $\phi$ is strictly decreasing, then $\min\ \phi(g(x))$ is equivalent to $\max\ g(x)$.

\paragraph{Constraints.}
For increasing $\phi$, the constraint $\phi(g(x)) \le \alpha$ is equivalent to
$g(x) \le \phi^{-1}(\alpha)$ (and similarly $\phi(g(x)) \ge \alpha \iff g(x) \ge \phi^{-1}(\alpha)$).
Example: $\log(y) \le \alpha \iff y \le e^{\alpha}$.

\paragraph{Notes.}
The bilinear case with one binary and one bounded continuous variable is exact with the four
McCormick inequalities above; if both variables are continuous, McCormick gives a relaxation.
For fractional objectives, positivity of $c^\top x + d$ on the feasible set is required.

\section{Technical results generation methodology}
\label{app:tech_details}
This section describes the detailed experimental methodology for evaluating the \textit{LinearizeLLM} framework with Google's \textit{Gemini 2.5 Flash} (\texttt{temperature}\,=\,0.05, \texttt{max tokens}\,=\,64000, \texttt{timeout}\,=\,120 s, \texttt{top\mbox{-}p}=0.9) across 40 optimization problems and three information scenarios.
\paragraph{Experimental Framework.}
Each experiment uses three fixed seeds (1, 2, 3) to ensure reproducibility. All experiments were conducted on a system with AMD Ryzen Threadripper PRO 7955WX processor @ 4.5 GHz, 512 GB RAM, and PNY NVIDIA T1000 8 GB graphics card. Detailed instructions for executing these experiments can be found in the project \texttt{README.md}.
\paragraph{Performance Evaluation.}
We evaluate against reference solves using OSR, DSR, RSR, and CSR. For each instance, we construct a reference formulation in \texttt{gurobipy} (hand-written for hand-crafted benchmarks; auto-generated for synthetic benchmarks) by applying Gurobi-supported exact reformulations  (including general constraints and bounded big-M for binary–continuous products, with required domain checks for linear-fractional constraints). We solve with Gurobi Optimizer 12 with default parameters and use the resulting objective value as the OSR reference.

\newpage

\section{Computational Efficiency and Token Analysis}\label{app:costs}
\begin{table}[ht]
\centering
\small
\label{tab:agent_performance}
\begin{tabular}{lllrr}
\toprule
\textbf{Depth} & \textbf{Method} & \textbf{Agent} & \textbf{Tokens} & \textbf{Avg. Time} \\
\midrule
\multirow{4}{*}{$d=1$} & LinearizeLLM & Pattern Detection & 6,219 $\pm$ 3,028 & 16.37s \\
                         & LinearizeLLM & Reformulation & 8,556 $\pm$ 3,975 & 27.54s \\
                         & LinearizeLLM & Code Generation & 4,792 $\pm$ 2,190 & 15.90s \\
                         \cmidrule{2-5}
                         & One-Shot     & One-Shot & 5,973 $\pm$ 2,409 & 21.88s \\
\midrule
\multirow{4}{*}{$d=2$} & LinearizeLLM & Pattern Detection & 14,037 $\pm$ 8,388 & 45.30s \\
                         & LinearizeLLM & Reformulation & 63,437 $\pm$ 49,267 & 193.64s \\
                         & LinearizeLLM & Code Generation & 11,401 $\pm$ 5,204 & 34.70s \\
                         \cmidrule{2-5}
                         & One-Shot     & One-Shot & 16,180 $\pm$ 5,274 & 59.74s \\
\midrule
\multirow{4}{*}{$d\geq3$} & LinearizeLLM & Pattern Detection & 14,486 $\pm$ 3,748 & 47.85s \\
                         & LinearizeLLM & Reformulation & 71,776 $\pm$ 41,938 & 222.59s \\
                         & LinearizeLLM & Code Generation & 9,274 $\pm$ 3,860 & 27.19s \\
                         \cmidrule{2-5}
                         & One-Shot     & One-Shot & 15,530 $\pm$ 10,793 & 56.74s \\
\bottomrule
\end{tabular}
\caption{Performance statistics per agent across nesting depths. Token counts are reported as mean $\pm$ standard deviation.}
\end{table}

\section{Perturbation Levels}\label{app:perturbations}
We define five levels of \LaTeX{} input perturbation, ranging from clean baseline models to highly modified versions that combine multiple noise types.

\begin{itemize}
  \item \textbf{L0 (Clean):} The original, unchanged \LaTeX{} optimization problems from our dataset.
  \item \textbf{L1 (Format Noise):} Introduces syntactic variations that do not change the mathematical symbols (e.g. arbitrary whitespace changes or inconsistent line breaks).
  \item \textbf{L2 (Macro + Refactor):} Introduces local macro definitions (e.g. using \texttt{\textbackslash def}) for variables and parameters, rewrites operator formatting (e.g., \texttt{\textbackslash max} vs. \texttt{max}), and splits long expressions into multiple lines.
  \item \textbf{L3 (Structure-preserving Reorder):} Modifies the logical structure of the model without altering its semantics. This includes reordering the sequence of constraints, swapping terms within summations, and switching between set notation (e.g., $i \in \mathcal{I}$) and index notation (e.g., $i=1, \dots, I$).
  \item \textbf{L4 (Combined):} A composite level that simultaneously applies all perturbations from L1, L2, and L3 to test the system's performance under extreme syntactic variance.
\end{itemize}

\subsection{Perturbation example L1}\label{app:l1}
\begin{Verbatim}[breaklines=true,breakanywhere=true]
\min_{x, z} \quad & \sum_{(i,j) \in \mathcal{L}} \sum_{p \in \mathcal{P}} \left( ShipmentCost_{i,j,p} * x_{i,j,p} - 0.2 * ShipmentCost_{i,j,p} * x_{i,j,p} * z_{i,j} \right) + z_{i,j} * ContractCosts_{i,j} \\
\text{s.t.} \quad & \sum_{j:\,(j,i)\in\mathcal{L}} x_{j,i,p} + Supply_{i,p} = \sum_{j:\,(i,j)\in\mathcal{L}} x_{i,j,p} + Demand_{i,p} && \forall i\in\mathcal{C}, p\in\mathcal{P}, \\
& x_{i,j,p} \leq Capacity_{i,j,p} && \forall (i,j)\in\mathcal{L}, p\in\mathcal{P}, \\
& \sum_{p\in\mathcal{P}} x_{i,j,p} \leq JointCapacity_{i,j} && \forall (i,j)\in\mathcal{L}, \\
& x_{i,j,p} \geq 0 && \forall (i,j)\in\mathcal{L}, p\in\mathcal{P} \\
& z_{i,j} \in \{0,1\} && \forall (i,j)\in\mathcal{L}
\end{Verbatim}

\subsection{Perturbation example L2}\label{app:l2}
\begin{Verbatim}[breaklines=true,breakanywhere=true]
\def\scost{ShipmentCost}
\def\ccost{ContractCosts}
\min_{x, z}\quad & \sum_{(i,j)\in\mathcal{L}}\sum_{p\in\mathcal{P}} \left(\scost_{i,j,p} \cdot x_{i,j,p} - (0.2 \cdot \scost_{i,j,p} \cdot x_{i,j,p} \cdot z_{i,j})\right) + (z_{i,j} \cdot \ccost_{i,j})\\
\text{s.t.}\quad
& \sum_{j:(j,i)\in\mathcal{L}} x_{j,i,p} + Supply_{i,p} = \sum_{j:(i,j)\in\mathcal{L}} x_{i,j,p} + Demand_{i,p}&& \forall i\in\mathcal{C}, p\in\mathcal{P},\\
& x_{i,j,p} \leq Capacity_{i,j,p} && \forall (i,j)\in\mathcal{L}, p\in\mathcal{P},\\
& \sum_{p\in\mathcal{P}} x_{i,j,p} \leq JointCapacity_{i,j} && \forall (i,j)\in\mathcal{L}, \\[6pt]
& x_{i,j,p} \geq 0 && \forall (i,j)\in\mathcal{L}, p\in\mathcal{P} \\[6pt]
& z_{i,j} \in \{0,1\} && \forall (i,j)\in\mathcal{L}
\end{Verbatim}

\subsection{Perturbation example L3}\label{app:l3}
\begin{Verbatim}[breaklines=true,breakanywhere=true]
\min_{x, z}\quad & \sum_{p\in\mathcal{P}}\sum_{(i,j)\in\mathcal{L}} \left(ShipmentCost_{i,j,p} * x_{i,j,p} - 0.2 * ShipmentCost_{i,j,p} * x_{i,j,p} * z_{i,j}\right) + z_{i,j} * ContractCosts_{i,j}\\
\text{s.t.}\quad
& x_{i,j,p} \;\leq\; Capacity_{i,j,p} && \forall (i,j)\in\mathcal{L},\;p\in\mathcal{P},\\
& \sum_{p\in\mathcal{P}} x_{i,j,p} \;\leq\; JointCapacity_{i,j} && \forall (i,j)\in\mathcal{L}, \\
& \sum_{j:\,(j,i)\in\mathcal{L}} x_{j,i,p} + Supply_{i,p} = \sum_{j:\,(i,j)\in\mathcal{L}} x_{i,j,p} + Demand_{i,p}&& \forall i\in\mathcal{C},\;p\in\mathcal{P},\\
& z_{i,j} \;\in\; \left\{0,1\right\} && \forall (i,j)\in\mathcal{L}, \\
& x_{i,j,p} \;\geq\; 0 && \forall (i,j)\in\mathcal{L},\;p\in\mathcal{P}
\end{Verbatim}

\subsection{Perturbation example L4}\label{app:l4}
\begin{Verbatim}[breaklines=true,breakanywhere=true]
\def\scost{ShipmentCost}
\def\ccost{ContractCosts}
\min_{x, z} \quad & \sum_{p \in \mathcal{P}} \sum_{(i,j) \in \mathcal{L}} \left( (\scost_{i,j,p} \cdot x_{i,j,p}) - (0.2 \cdot \scost_{i,j,p} \cdot x_{i,j,p} \cdot z_{i,j}) \right) + (z_{i,j} \cdot \ccost_{i,j}) \\
\text{s.t.} \quad & x_{i,j,p} \leq Capacity_{i,j,p} && \forall (i,j) \in \mathcal{L}, p \in \mathcal{P}, \\
& \sum_{p \in \mathcal{P}} x_{i,j,p} \leq JointCapacity_{i,j} && \forall (i,j) \in \mathcal{L}, \\
& \sum_{j:(j,i) \in \mathcal{L}} x_{j,i,p} + Supply_{i,p} = \sum_{j:(i,j) \in \mathcal{L}} x_{i,j,p} + Demand_{i,p} && \forall i \in \mathcal{C}, p \in \mathcal{P}, \\
& z_{i,j} \in \{0,1\} && \forall (i,j) \in \mathcal{L}, \\
& x_{i,j,p} \geq 0 && \forall (i,j) \in \mathcal{L}, p \in \mathcal{P}
\end{Verbatim}

\section{Context ablation levels}
\label{app:context_ablation}
We define six incremental context categories used in the context ablation study (Section~\ref{sec:context_ablation}):

\begin{itemize}
  \item \textbf{C1 (Minimal):} Provides only the mathematical model with no external metadata; agents receive only the component type (objective vs.\ constraint) for each term.
  \item \textbf{C2 (Variables Only):} Adds decision variable definitions.
  \item \textbf{C3 (Parameters Only):} Provides only the names of concrete parameters to distinguish them from decision variables.
  \item \textbf{C4 (Vars + Param Names):} Combines metadata from C2 and C3.
  \item \textbf{C5 (Vars + Param Values):} Extends C4 by including the concrete numerical values of all parameters.
  \item \textbf{C6 (Full Context):} Includes all previous metadata plus the full local constraint data.
\end{itemize}

We evaluate these context levels on 10 randomly sampled instances (5 with $d=2$ and 5 with $d\ge 3$).

\section{Arg min comparison for original vs.\ linearized models}\label{app:depth_1_results}
We report side-by-side optimal solutions (objective value and variable assignments) for the original nonlinear model as additional supporting evidence for equivalent reformulations beyond OSR. Since Gurobi and LinearizeLLM introduce auxiliary variables and may use different variable names, comparisons should be made after accounting for these naming differences and auxiliaries. For conciseness, we show only the first run per depth-1 problem, regardless of whether the run satisfies the OSR criterion; the remaining runs are provided in the supplementary code.


\subsection{problem\_blend\_problem\_abs}

\begin{minipage}[t]{0.48\linewidth}
\centering
\textbf{Original}\par\medskip
\begin{tabular}{@{}l r@{}}
\hline
\textbf{Objective} & $10.0$ \\
\hline
\textbf{Variable} & \textbf{Value} \\
\hline
\texttt{absolute\_deviations.0} & $0.0$ \\
\texttt{absolute\_deviations.1} & $0.0$ \\
\texttt{alloy\_amounts.0} & $1.0$ \\
\texttt{alloy\_amounts.1} & $0.0$ \\
\texttt{signed\_deviations.0} & $0.0$ \\
\texttt{signed\_deviations.1} & $0.0$ \\
\hline
\end{tabular}
\end{minipage}
\hfill
\begin{minipage}[t]{0.48\linewidth}
\centering
\textbf{LinearizeLLM}\par\medskip
\begin{tabular}{@{}l r@{}}
\hline
\textbf{Objective} & $10.0$ \\
\hline
\textbf{Variable} & \textbf{Value} \\
\hline
\texttt{x[0]} & $1.0$ \\
\texttt{x[1]} & $0.0$ \\
\texttt{z[0]} & $0.0$ \\
\texttt{z[1]} & $0.0$ \\
\hline
\end{tabular}
\end{minipage}

\subsection{problem\_blend\_problem\_frac}

\begin{minipage}[t]{0.48\linewidth}
\centering
\textbf{Original}\par\medskip
\begin{tabular}{@{}l r@{}}
\hline
\textbf{Objective} & $0.05749999999999999$ \\
\hline
\textbf{Variable} & \textbf{Value} \\
\hline
\texttt{cc\_variables.0} & $0.0$ \\
\texttt{cc\_variables.1} & $0.027777777777777693$ \\
\texttt{cc\_variables.2} & $0.1111111111111112$ \\
\texttt{original\_mix.0} & $0.0$ \\
\texttt{original\_mix.1} & $199.9999999999994$ \\
\texttt{original\_mix.2} & $800.0000000000007$ \\
\texttt{ratio\_variable} & $0.0001388888888888889$ \\
\hline
\end{tabular}
\end{minipage}
\hfill
\begin{minipage}[t]{0.48\linewidth}
\centering
\textbf{LinearizeLLM}\par\medskip
\begin{tabular}{@{}l r@{}}
\hline
\textbf{Objective} & $0.0$ \\
\hline
\textbf{Variable} & \textbf{Value} \\
\hline
\texttt{t} & $0.0$ \\
\texttt{y[0]} & $0.0$ \\
\texttt{y[1]} & $0.0$ \\
\texttt{y[2]} & $0.0$ \\
\hline
\end{tabular}
\end{minipage}

\subsection{problem\_diet\_problem\_min\_abs}

\begin{minipage}[t]{0.48\linewidth}
\centering
\textbf{Original}\par\medskip
\begin{tabular}{@{}l r@{}}
\hline
\textbf{Objective} & $9.333333333333334$ \\
\hline
\textbf{Variable} & \textbf{Value} \\
\hline
\texttt{absolute\_cost\_gap} & $9.333333333333334$ \\
\texttt{cost\_difference} & $-9.333333333333334$ \\
\texttt{food\_quantities.0} & $9.333333333333334$ \\
\texttt{food\_quantities.1} & $1.3333333333333333$ \\
\texttt{min\_values.0} & $50.0$ \\
\texttt{min\_values.1} & $30.0$ \\
\texttt{nutrient\_intake.0} & $100.0$ \\
\texttt{nutrient\_intake.1} & $60.0$ \\
\hline
\end{tabular}
\end{minipage}
\hfill
\begin{minipage}[t]{0.48\linewidth}
\centering
\textbf{LinearizeLLM}\par\medskip
\begin{tabular}{@{}l r@{}}
\hline
\textbf{Objective} & $9.333333333333334$ \\
\hline
\textbf{Variable} & \textbf{Value} \\
\hline
\texttt{b[0]} & $0.0$ \\
\texttt{b[1]} & $1.0$ \\
\texttt{x[0]} & $9.333333333333334$ \\
\texttt{x[1]} & $1.3333333333333333$ \\
\texttt{z\_abs\_1} & $9.333333333333334$ \\
\hline
\end{tabular}
\end{minipage}

\subsection{problem\_diet\_problem\_monotone}

\begin{minipage}[t]{0.48\linewidth}
\centering
\textbf{Original}\par\medskip
\begin{tabular}{@{}l r@{}}
\hline
\textbf{Objective} & $30740.40934396602$ \\
\hline
\textbf{Variable} & \textbf{Value} \\
\hline
\texttt{0} & $4.666666666666667$ \\
\texttt{1} & $0.6666666666666664$ \\
\hline
\end{tabular}
\end{minipage}
\hfill
\begin{minipage}[t]{0.48\linewidth}
\centering
\textbf{LinearizeLLM}\par\medskip
\begin{tabular}{@{}l r@{}}
\hline
\textbf{Objective} & $30740.40934396602$ \\
\hline
\textbf{Variable} & \textbf{Value} \\
\hline
\texttt{x[0]} & $4.666666666666667$ \\
\texttt{x[1]} & $0.6666666666666664$ \\
\hline
\end{tabular}
\end{minipage}

\subsection{problem\_diet\_problem\_nonlinear\_frac}

\begin{minipage}[t]{0.48\linewidth}
\centering
\textbf{Original}\par\medskip
\begin{tabular}{@{}l r@{}}
\hline
\textbf{Objective} & $1.5$ \\
\hline
\textbf{Variable} & \textbf{Value} \\
\hline
\texttt{cost\_ratio} & $1.5$ \\
\texttt{food\_quantities.0} & $0.0$ \\
\texttt{food\_quantities.1} & $3.6698321894119537$ \\
\hline
\end{tabular}
\end{minipage}
\hfill
\begin{minipage}[t]{0.48\linewidth}
\centering
\textbf{LinearizeLLM}\par\medskip
\begin{tabular}{@{}l r@{}}
\hline
\textbf{Objective} & $1.5$ \\
\hline
\textbf{Variable} & \textbf{Value} \\
\hline
\texttt{t} & $1.0$ \\
\texttt{y\_1} & $0.0$ \\
\texttt{y\_2} & $1.0$ \\
\hline
\end{tabular}
\end{minipage}

\subsection{problem\_knapsack\_problem\_nonlinear\_min\_1}

\begin{minipage}[t]{0.48\linewidth}
\centering
\textbf{Original}\par\medskip
\begin{tabular}{@{}l r@{}}
\hline
\textbf{Objective} & $280.0$ \\
\hline
\textbf{Variable} & \textbf{Value} \\
\hline
\texttt{item\_selections.0} & $1.0$ \\
\texttt{item\_selections.1} & $1.0$ \\
\texttt{item\_selections.2} & $1.0$ \\
\texttt{min\_x1\_x3} & $1.0$ \\
\hline
\end{tabular}
\end{minipage}
\hfill
\begin{minipage}[t]{0.48\linewidth}
\centering
\textbf{LinearizeLLM}\par\medskip
\begin{tabular}{@{}l r@{}}
\hline
\textbf{Objective} & $280.0$ \\
\hline
\textbf{Variable} & \textbf{Value} \\
\hline
\texttt{aux\_1} & $1.0$ \\
\texttt{x[0]} & $1.0$ \\
\texttt{x[1]} & $1.0$ \\
\texttt{x[2]} & $1.0$ \\
\hline
\end{tabular}
\end{minipage}

\subsection{problem\_knapsack\_problem\_nonlinear\_min\_2}

\begin{minipage}[t]{0.48\linewidth}
\centering
\textbf{Original}\par\medskip
\begin{tabular}{@{}l r@{}}
\hline
\textbf{Objective} & $220.0$ \\
\hline
\textbf{Variable} & \textbf{Value} \\
\hline
\texttt{item\_selections.0} & $-0.0$ \\
\texttt{item\_selections.1} & $1.0$ \\
\texttt{item\_selections.2} & $1.0$ \\
\texttt{min\_x0\_x1} & $0.0$ \\
\hline
\end{tabular}
\end{minipage}
\hfill
\begin{minipage}[t]{0.48\linewidth}
\centering
\textbf{LinearizeLLM}\par\medskip
\begin{tabular}{@{}l r@{}}
\hline
\textbf{Objective} & $220.0$ \\
\hline
\textbf{Variable} & \textbf{Value} \\
\hline
\texttt{aux\_1} & $0.0$ \\
\texttt{x[0]} & $-0.0$ \\
\texttt{x[1]} & $1.0$ \\
\texttt{x[2]} & $1.0$ \\
\hline
\end{tabular}
\end{minipage}

\subsection{problem\_media\_selection\_nonlinear\_binbin}

\begin{minipage}[t]{0.48\linewidth}
\centering
\textbf{Original}\par\medskip
\begin{tabular}{@{}l r@{}}
\hline
\textbf{Objective} & $20.0$ \\
\hline
\textbf{Variable} & \textbf{Value} \\
\hline
\texttt{media\_selections.0} & $1.0$ \\
\texttt{media\_selections.1} & $1.0$ \\
\texttt{media\_selections.2} & $0.0$ \\
\hline
\end{tabular}
\end{minipage}
\hfill
\begin{minipage}[t]{0.48\linewidth}
\centering
\textbf{LinearizeLLM}\par\medskip
\begin{tabular}{@{}l r@{}}
\hline
\textbf{Objective} & $20.0$ \\
\hline
\textbf{Variable} & \textbf{Value} \\
\hline
\texttt{aux\_1} & $1.0$ \\
\texttt{x[0]} & $1.0$ \\
\texttt{x[1]} & $1.0$ \\
\texttt{x[2]} & $0.0$ \\
\hline
\end{tabular}
\end{minipage}

\subsection{problem\_media\_selection\_nonlinear\_bincon}

\begin{minipage}[t]{0.48\linewidth}
\centering
\textbf{Original}\par\medskip
\begin{tabular}{@{}l r@{}}
\hline
\textbf{Objective} & $15.0$ \\
\hline
\textbf{Variable} & \textbf{Value} \\
\hline
\texttt{media\_selections.0} & $1.0$ \\
\texttt{media\_selections.1} & $1.0$ \\
\texttt{media\_selections.2} & $0.0$ \\
\hline
\end{tabular}
\end{minipage}
\hfill
\begin{minipage}[t]{0.48\linewidth}
\centering
\textbf{LinearizeLLM}\par\medskip
\begin{tabular}{@{}l r@{}}
\hline
\textbf{Objective} & $15.0$ \\
\hline
\textbf{Variable} & \textbf{Value} \\
\hline
\texttt{q[0]} & $0.0$ \\
\texttt{q[1]} & $5.0$ \\
\texttt{q[2]} & $0.0$ \\
\texttt{x[0]} & $1.0$ \\
\texttt{x[1]} & $1.0$ \\
\texttt{x[2]} & $-0.0$ \\
\texttt{y[0]} & $0.0$ \\
\texttt{y[1]} & $5.0$ \\
\texttt{y[2]} & $0.0$ \\
\hline
\end{tabular}
\end{minipage}

\subsection{problem\_multi\_nonlinear\_abs}

\begin{minipage}[t]{0.48\linewidth}
\centering
\textbf{Original}\par\medskip
\begin{tabular}{@{}l r@{}}
\hline
\textbf{Objective} & $240.0$ \\
\hline
\textbf{Variable} & \textbf{Value} \\
\hline
\texttt{deviations.(0, 0)} & $0.0$ \\
\texttt{deviations.(0, 1)} & $5.0$ \\
\texttt{deviations.(1, 0)} & $0.0$ \\
\texttt{deviations.(1, 1)} & $5.0$ \\
\texttt{differences.(0, 0)} & $0.0$ \\
\texttt{differences.(0, 1)} & $-5.0$ \\
\texttt{differences.(1, 0)} & $0.0$ \\
\texttt{differences.(1, 1)} & $5.0$ \\
\texttt{shipments.(0, 0, 0)} & $20.0$ \\
\texttt{shipments.(0, 0, 1)} & $15.0$ \\
\texttt{shipments.(0, 1, 0)} & $0.0$ \\
\texttt{shipments.(0, 1, 1)} & $15.0$ \\
\texttt{shipments.(1, 0, 0)} & $10.0$ \\
\texttt{shipments.(1, 0, 1)} & $10.0$ \\
\texttt{shipments.(1, 1, 0)} & $30.0$ \\
\texttt{shipments.(1, 1, 1)} & $0.0$ \\
\hline
\end{tabular}
\end{minipage}
\hfill
\begin{minipage}[t]{0.48\linewidth}
\centering
\textbf{LinearizeLLM}\par\medskip
\begin{tabular}{@{}l r@{}}
\hline
\textbf{Objective} & $240.0$ \\
\hline
\textbf{Variable} & \textbf{Value} \\
\hline
\texttt{aux[0,0]} & $0.0$ \\
\texttt{aux[0,1]} & $5.0$ \\
\texttt{aux[1,0]} & $0.0$ \\
\texttt{aux[1,1]} & $5.0$ \\
\texttt{x[0,0,0]} & $20.0$ \\
\texttt{x[0,0,1]} & $15.0$ \\
\texttt{x[0,1,0]} & $0.0$ \\
\texttt{x[0,1,1]} & $15.0$ \\
\texttt{x[1,0,0]} & $10.0$ \\
\texttt{x[1,0,1]} & $10.0$ \\
\texttt{x[1,1,0]} & $30.0$ \\
\texttt{x[1,1,1]} & $0.0$ \\
\hline
\end{tabular}
\end{minipage}

\subsection{problem\_netasgn\_nonlinear\_abs}

\begin{minipage}[t]{0.48\linewidth}
\centering
\textbf{Original}\par\medskip
\begin{tabular}{@{}l r@{}}
\hline
\textbf{Objective} & $297.0$ \\
\hline
\textbf{Variable} & \textbf{Value} \\
\hline
\texttt{assignments.(0, 0)} & $2.0$ \\
\texttt{assignments.(0, 1)} & $6.0$ \\
\texttt{assignments.(1, 0)} & $3.0$ \\
\texttt{assignments.(1, 1)} & $4.0$ \\
\texttt{deviation\_pairs.(0, 1)} & $1.0$ \\
\texttt{differences.(0, 1)} & $1.0$ \\
\texttt{max\_deviation} & $1.0$ \\
\hline
\end{tabular}
\end{minipage}
\hfill
\begin{minipage}[t]{0.48\linewidth}
\centering
\textbf{LinearizeLLM}\par\medskip
\begin{tabular}{@{}l r@{}}
\hline
\textbf{Objective} & $297.0$ \\
\hline
\textbf{Variable} & \textbf{Value} \\
\hline
\texttt{aux\_1} & $1.0$ \\
\texttt{x[0,0]} & $2.0$ \\
\texttt{x[0,1]} & $6.0$ \\
\texttt{x[1,0]} & $3.0$ \\
\texttt{x[1,1]} & $4.0$ \\
\hline
\end{tabular}
\end{minipage}

\subsection{problem\_netasgn\_nonlinear\_max}

\begin{minipage}[t]{0.48\linewidth}
\centering
\textbf{Original}\par\medskip
\begin{tabular}{@{}l r@{}}
\hline
\textbf{Objective} & $286.0$ \\
\hline
\textbf{Variable} & \textbf{Value} \\
\hline
\texttt{assignments.(0, 0)} & $2.0$ \\
\texttt{assignments.(0, 1)} & $6.0$ \\
\texttt{assignments.(1, 0)} & $3.0$ \\
\texttt{assignments.(1, 1)} & $4.0$ \\
\texttt{deviations.dev\_0\_1} & $1.0$ \\
\texttt{deviations.diff\_0\_1} & $1.0$ \\
\texttt{max\_deviation} & $1.0$ \\
\texttt{total\_hours.0} & $8.0$ \\
\texttt{total\_hours.1} & $7.0$ \\
\hline
\end{tabular}
\end{minipage}
\hfill
\begin{minipage}[t]{0.48\linewidth}
\centering
\textbf{LinearizeLLM}\par\medskip
\begin{tabular}{@{}l r@{}}
\hline
\textbf{Objective} & $297.0$ \\
\hline
\textbf{Variable} & \textbf{Value} \\
\hline
\texttt{aux\_1} & $1.0$ \\
\texttt{x[0,0]} & $2.0$ \\
\texttt{x[0,1]} & $6.0$ \\
\texttt{x[1,0]} & $3.0$ \\
\texttt{x[1,1]} & $4.0$ \\
\hline
\end{tabular}
\end{minipage}

\subsection{problem\_netmcol\_nonlinear\_bincon}

\begin{minipage}[t]{0.48\linewidth}
\centering
\textbf{Original}\par\medskip
\begin{tabular}{@{}l r@{}}
\hline
\textbf{Objective} & $10.0$ \\
\hline
\textbf{Variable} & \textbf{Value} \\
\hline
\texttt{contracts.(0, 1)} & $0.0$ \\
\texttt{contracts.(1, 0)} & $0.0$ \\
\texttt{shipments.(0, 1, 0)} & $10.0$ \\
\texttt{shipments.(1, 0, 0)} & $0.0$ \\
\hline
\end{tabular}
\end{minipage}
\hfill
\begin{minipage}[t]{0.48\linewidth}
\centering
\textbf{LinearizeLLM}\par\medskip
\begin{tabular}{@{}l r@{}}
\hline
\textbf{Objective} & $10.0$ \\
\hline
\textbf{Variable} & \textbf{Value} \\
\hline
\texttt{aux\_1[0,0,0]} & $0.0$ \\
\texttt{aux\_1[0,1,0]} & $0.0$ \\
\texttt{aux\_1[1,0,0]} & $0.0$ \\
\texttt{aux\_1[1,1,0]} & $0.0$ \\
\texttt{x[0,0,0]} & $0.0$ \\
\texttt{x[0,1,0]} & $10.0$ \\
\texttt{x[1,0,0]} & $0.0$ \\
\texttt{x[1,1,0]} & $0.0$ \\
\texttt{z[0,0]} & $0.0$ \\
\texttt{z[0,1]} & $0.0$ \\
\texttt{z[1,0]} & $0.0$ \\
\texttt{z[1,1]} & $0.0$ \\
\hline
\end{tabular}
\end{minipage}

\subsection{problem\_netmcol\_nonlinear\_frac}

\begin{minipage}[t]{0.48\linewidth}
\centering
\textbf{Original}\par\medskip
\begin{tabular}{@{}l r@{}}
\hline
\textbf{Objective} & $0.2$ \\
\hline
\textbf{Variable} & \textbf{Value} \\
\hline
\texttt{demand.(0, 0)} & $0.0$ \\
\texttt{demand.(1, 0)} & $10.0$ \\
\texttt{denominator} & $10.0$ \\
\texttt{numerator} & $2.0$ \\
\texttt{ratio} & $0.2$ \\
\texttt{shipments.(0, 0)} & $10.0$ \\
\hline
\end{tabular}
\end{minipage}
\hfill
\begin{minipage}[t]{0.48\linewidth}
\centering
\textbf{LinearizeLLM}\par\medskip
\begin{tabular}{@{}l r@{}}
\hline
\textbf{Objective} & $0.19980019980019983$ \\
\hline
\textbf{Variable} & \textbf{Value} \\
\hline
\texttt{t} & $0.09990009990009992$ \\
\texttt{x\_hat[0,0]} & $0.9990009990009991$ \\
\texttt{y\_hat[0,0]} & $0.0$ \\
\texttt{y\_hat[1,0]} & $0.9990009990009991$ \\
\hline
\end{tabular}
\end{minipage}

\subsection{problem\_nltrans\_nonlinear\_bincon}

\begin{minipage}[t]{0.48\linewidth}
\centering
\textbf{Original}\par\medskip
\begin{tabular}{@{}l r@{}}
\hline
\textbf{Objective} & $305.0$ \\
\hline
\textbf{Variable} & \textbf{Value} \\
\hline
\texttt{investment} & $0.0$ \\
\texttt{shipments.(0, 0)} & $5.0$ \\
\texttt{shipments.(0, 1)} & $15.0$ \\
\texttt{shipments.(1, 0)} & $25.0$ \\
\texttt{shipments.(1, 1)} & $5.0$ \\
\hline
\end{tabular}
\end{minipage}
\hfill
\begin{minipage}[t]{0.48\linewidth}
\centering
\textbf{LinearizeLLM}\par\medskip
\begin{tabular}{@{}l r@{}}
\hline
\textbf{Objective} & $305.0$ \\
\hline
\textbf{Variable} & \textbf{Value} \\
\hline
\texttt{aux1} & $305.0$ \\
\texttt{aux2} & $0.0$ \\
\texttt{x[0,0]} & $5.0$ \\
\texttt{x[0,1]} & $15.0$ \\
\texttt{x[1,0]} & $25.0$ \\
\texttt{x[1,1]} & $5.0$ \\
\texttt{z} & $0.0$ \\
\hline
\end{tabular}
\end{minipage}

\subsection{problem\_nltrans\_nonlinear\_max}

\begin{minipage}[t]{0.48\linewidth}
\centering
\textbf{Original}\par\medskip
\begin{tabular}{@{}l r@{}}
\hline
\textbf{Objective} & $305.0$ \\
\hline
\textbf{Variable} & \textbf{Value} \\
\hline
\texttt{differences.(0, 0)} & $-10.0$ \\
\texttt{differences.(0, 1)} & $-10.0$ \\
\texttt{differences.(1, 0)} & $0.0$ \\
\texttt{differences.(1, 1)} & $-15.0$ \\
\texttt{excess.(0, 0)} & $0.0$ \\
\texttt{excess.(0, 1)} & $0.0$ \\
\texttt{excess.(1, 0)} & $0.0$ \\
\texttt{excess.(1, 1)} & $0.0$ \\
\texttt{shipments.(0, 0)} & $5.0$ \\
\texttt{shipments.(0, 1)} & $15.0$ \\
\texttt{shipments.(1, 0)} & $25.0$ \\
\texttt{shipments.(1, 1)} & $5.0$ \\
\hline
\end{tabular}
\end{minipage}
\hfill
\begin{minipage}[t]{0.48\linewidth}
\centering
\textbf{LinearizeLLM}\par\medskip
\begin{tabular}{@{}l r@{}}
\hline
\textbf{Objective} & $305.0$ \\
\hline
\textbf{Variable} & \textbf{Value} \\
\hline
\texttt{x[0,0]} & $5.0$ \\
\texttt{x[0,1]} & $15.0$ \\
\texttt{x[1,0]} & $25.0$ \\
\texttt{x[1,1]} & $5.0$ \\
\texttt{y[0,0]} & $0.0$ \\
\texttt{y[0,1]} & $0.0$ \\
\texttt{y[1,0]} & $0.0$ \\
\texttt{y[1,1]} & $0.0$ \\
\hline
\end{tabular}
\end{minipage}

\subsection{problem\_prod\_nonlinear\_bincon}

\begin{minipage}[t]{0.48\linewidth}
\centering
\textbf{Original}\par\medskip
\begin{tabular}{@{}l r@{}}
\hline
\textbf{Objective} & $55.66666666666667$ \\
\hline
\textbf{Variable} & \textbf{Value} \\
\hline
\texttt{campaign} & $0.0$ \\
\texttt{production.0} & $4.0$ \\
\texttt{production.1} & $1.1666666666666667$ \\
\texttt{production.2} & $3.0$ \\
\hline
\end{tabular}
\end{minipage}
\hfill
\begin{minipage}[t]{0.48\linewidth}
\centering
\textbf{LinearizeLLM}\par\medskip
\begin{tabular}{@{}l r@{}}
\hline
\textbf{Objective} & $55.66666666666667$ \\
\hline
\textbf{Variable} & \textbf{Value} \\
\hline
\texttt{aux\_1} & $55.66666666666667$ \\
\texttt{aux\_2} & $0.0$ \\
\texttt{x[0]} & $4.0$ \\
\texttt{x[1]} & $1.1666666666666667$ \\
\texttt{x[2]} & $3.0$ \\
\texttt{z} & $0.0$ \\
\hline
\end{tabular}
\end{minipage}

\subsection{problem\_prod\_nonlinear\_max}

\begin{minipage}[t]{0.48\linewidth}
\centering
\textbf{Original}\par\medskip
\begin{tabular}{@{}l r@{}}
\hline
\textbf{Objective} & $89.5$ \\
\hline
\textbf{Variable} & \textbf{Value} \\
\hline
\texttt{consumption\_minus\_b} & $4.833333333333333$ \\
\texttt{excess} & $4.833333333333333$ \\
\texttt{production.0} & $4.0$ \\
\texttt{production.1} & $6.0$ \\
\texttt{production.2} & $3.0$ \\
\hline
\end{tabular}
\end{minipage}
\hfill
\begin{minipage}[t]{0.48\linewidth}
\centering
\textbf{LinearizeLLM}\par\medskip
\begin{tabular}{@{}l r@{}}
\hline
\textbf{Objective} & $89.5$ \\
\hline
\textbf{Variable} & \textbf{Value} \\
\hline
\texttt{aux\_1} & $4.833333333333333$ \\
\texttt{x[0]} & $4.0$ \\
\texttt{x[1]} & $6.0$ \\
\texttt{x[2]} & $3.0$ \\
\hline
\end{tabular}
\end{minipage}

\subsection{problem\_revenue\_maximization\_nonlinear\_bincon}

\begin{minipage}[t]{0.48\linewidth}
\centering
\textbf{Original}\par\medskip
\begin{tabular}{@{}l r@{}}
\hline
\textbf{Objective} & $8000.0$ \\
\hline
\textbf{Variable} & \textbf{Value} \\
\hline
\texttt{campaign} & $1.0$ \\
\texttt{package\_sales.0} & $20.0$ \\
\texttt{package\_sales.1} & $40.0$ \\
\hline
\end{tabular}
\end{minipage}
\hfill
\begin{minipage}[t]{0.48\linewidth}
\centering
\textbf{LinearizeLLM}\par\medskip
\begin{tabular}{@{}l r@{}}
\hline
\textbf{Objective} & $8000.0$ \\
\hline
\textbf{Variable} & \textbf{Value} \\
\hline
\texttt{aux\_1} & $0.0$ \\
\texttt{aux\_2} & $8800.0$ \\
\texttt{x[0]} & $20.0$ \\
\texttt{x[1]} & $40.0$ \\
\texttt{z} & $1.0$ \\
\hline
\end{tabular}
\end{minipage}


\end{document}